\PassOptionsToPackage{dvipsnames,table}{xcolor}
\documentclass[]{pediamed-ai}

\title{Evaluating Cognitive Age Alignment in Interactive AI Agents}

\author[1,2,*]{Yifan Shen}
\author[2,4,*]{Jiawen Zhang}
\author[2]{Jian Xu}
\author[2]{Junho Kim}
\author[2]{Ismini Lourentzou}
\author[1,2,\dagger]{Xu Cao}
\author[1,3,5,\dagger]{Meihuan Huang}

\affiliation[1]{PediaMed AI}
\affiliation[2]{University of Illinois Urbana-Champaign}
\affiliation[3]{Shenzhen Children's Hospital}
\affiliation[4]{Peking University}
\affiliation[5]{Hong Kong Polytechnic University}

\contribution[*]{Equal contribution}
\contribution[\dagger]{Corresponding author}

\newcolumntype{L}[1]{>{\raggedright\let\newline\\\arraybackslash\hspace{0pt}}m{#1}}
\newcolumntype{R}[1]{>{\raggedleft\let\newline\\\arraybackslash\hspace{0pt}}m{#1}}

\newcommand{\ignore}[1]{}

\makeatletter
\DeclareRobustCommand\onedot{\futurelet\@let@token\@onedot}
\def\@onedot{\ifx\@let@token.\else.\null\fi\xspace}

\makeatother

\definecolor{MyBlue}{rgb}{0.46, 0.50, 0.61}
\definecolor{MyDarkBlue}{rgb}{0,0.08,0.8}
\definecolor{MyDarkGreen}{RGB}{45,155,45}
\definecolor{MyDarkRed}{rgb}{0.8,0.02,0.02}
\definecolor{MyOrange}{rgb}{1.0, 0.4, 0.2}
\definecolor{MyPurple}{RGB}{111,0,255}
\definecolor{MyRed}{rgb}{0.8,0.0,0.0}
\definecolor{MyGold}{rgb}{0.75,0.6,0.12}
\definecolor{MyDarkgray}{rgb}{0.66, 0.66, 0.66}
\definecolor{MyBrown}{rgb}{0.65, 0.16, 0.16}
\definecolor{MyMutedRose}{rgb}{0.58, 0.29, 0.35}
\definecolor{JiayuanColor}{rgb}{0.60,0.43,0.48}
\definecolor{erranColor}{rgb}{24, 40, 113}

\definecolor{citecolor}{HTML}{696FAD}

\definecolor{bggray}{HTML}{F5F5F5}
\definecolor{pvdblue}{HTML}{DAE8FC}
\definecolor{RoseQuartzBg}{HTML}{F7CAC9}
\definecolor{RoseQuartz}{HTML}{F5A798}
\definecolor{Serenity}{HTML}{92A8D1}
\definecolor{OrangeRed}{rgb}{1.0, 0.27, 0.0}
\definecolor{RoyalBlue}{cmyk}{1, 0.50, 0, 0}
\definecolor{Turquoise}{HTML}{0F4C81}
\definecolor{mint}{rgb}{0.24, 0.71, 0.54}
\definecolor{green}{rgb}{0.0, 0.120, 0.0}

\newdimen\abovecrulesep
\newdimen\belowcrulesep
\makeatletter
\patchcmd{\@@@cmidrule}{\aboverulesep}{\abovecrulesep}{}{}
\patchcmd{\@xcmidrule}{\belowrulesep}{\belowcrulesep}{}{}
\makeatother

\definecolor{mybluetitle}{HTML}{4B527E} %

\definecolor{codegreen}{HTML}{478058}%
\definecolor{codegray}{rgb}{0.5,0.5,0.5}
\definecolor{codepurple}{HTML}{4F5E80} %
\definecolor{backcolour}{rgb}{0.95,0.95,0.92}
\lstdefinestyle{mystyle}{
    backgroundcolor=\color{backcolour},
    commentstyle=\color{codegreen},
    keywordstyle=\color{magenta},
    numberstyle=\tiny\color{codegray},
    stringstyle=\color{codepurple},
    basicstyle=\ttfamily\scriptsize,
    breakatwhitespace=false,
    breaklines=true,
    captionpos=b,
    keepspaces=true,
    frame=none,
    numbersep=5pt,
    showspaces=false,
    showstringspaces=false,
    showtabs=false,
    tabsize=2
}

\newtcolorbox{promptbox}[2][]{
    enhanced, 
    breakable,
    center title,
    left*=0pt, right*=0pt,
    boxsep=2pt, left=5pt, right=5pt,
    skin first=enhanced,
    skin middle=enhanced,
    skin last=enhanced,
    colback  = backcolour,
    fonttitle=\bfseries\rmfamily,
    fontupper=\scriptsize,
    title={\footnotesize\strut{#2}},
    #1
    }

\newtcolorbox{onebox}[2][]{
    enhanced, 
    center title,
    left*=0pt, right*=0pt,
    boxsep=2pt, left=5pt, right=5pt,
    skin first=enhanced,
    skin middle=enhanced,
    skin last=enhanced,
    colframe = mybluetitle!90,
  colback  = mybluetitle!10,
    fonttitle=\bfseries\rmfamily\fontfamily{phv}\selectfont,
    title={\footnotesize\strut{#2}  \refstepcounter{subsubsection} \addcontentsline{toc}{subsubsection}{\string\numberline{\thesubsubsection}#2}
    },
    #1
    }

\usepackage{amsmath,amsfonts,bm}

\def\eqref#1{equation~\ref{#1}}

\def\1{\bm{1}}

\DeclareMathAlphabet{\mathsfit}{\encodingdefault}{\sfdefault}{m}{sl}
\SetMathAlphabet{\mathsfit}{bold}{\encodingdefault}{\sfdefault}{bx}{n}

\usepackage{color}
\usepackage{epsfig}
\usepackage{graphicx}

\usepackage{booktabs}
\usepackage{tabularx}
\usepackage{ltablex}
\usepackage{tabularray}
\newcolumntype{C}{>{\centering\arraybackslash}X}
\usepackage{multirow}
\usepackage{diagbox}
\usepackage{hhline}

\usepackage{extarrows}
\usepackage{makecell}
\usepackage{wrapfig}
\usepackage{colortbl}
\usepackage{longtable}
\usepackage{fancyvrb}
\usepackage{listings}

\usepackage{adjustbox}
\usepackage{array}

\usepackage{amsmath,amsfonts,amssymb}
\usepackage{bm}
\usepackage{nicefrac}
\usepackage{microtype}

\usepackage{changepage}
\usepackage{extramarks}
\usepackage{fancyhdr}
\usepackage{setspace}
\usepackage{soul}
\usepackage{xspace}
\usepackage{multicol}

\usepackage{url}

\usepackage{enumerate}
\usepackage{enumitem}
\setlist[itemize]{leftmargin=*}

\usepackage{pifont}

\usepackage{algorithm,algpseudocode}

\usepackage[symbol]{footmisc}

\usepackage[most]{tcolorbox}

\usepackage{caption}
\usepackage{scalefnt}
\usepackage{fontawesome5}

\usepackage{titletoc}
\definecolor{citecolor}{HTML}{0071BC}
\definecolor{linkcolor}{HTML}{ED1C24}
\usepackage{hyperref}

\usepackage{marginnote}

\usepackage{lipsum}
\usepackage{nicematrix}
\usepackage{subcaption}

\usepackage{siunitx}
\usepackage{calc}
\usepackage{cleveref}
\crefname{appendix}{Appendix}{Appendices}
\crefname{section}{Section}{Sections}
\crefname{figure}{Fig.}{Figs.}
\crefname{table}{Tab.}{Tabs.}

\usepackage{titlecaps}
\usepackage{wasysym}

\usepackage{etoolbox}

\definecolor{ChildC}{HTML}{2F6DB3}   
\definecolor{ChildA}{HTML}{6A4FB3}   
\definecolor{ChildE}{HTML}{C05A9D}   

\newcommand{\benchmarknamenc}{%
 {\textsc{%
{Child}{Agent}{Eval}}}}

\newcommand{\kindatiny}{\fontsize{6pt}{7.2pt}\selectfont}
\newlength\savewidth

\definecolor{qcolor}{HTML}{536872}

\newcolumntype{P}[1]{>{\centering\arraybackslash}p{#1}}

\newcommand{\tablestyle}[2]{%
	\fontfamily{ptm}\selectfont%
	\let\itold\it%
	\def\it{\itold \fontfamily{ptm}\selectfont}%
	\setlength{\tabcolsep}{#1}\renewcommand{\arraystretch}{#2}\centering\kindatiny%
	\let\citeold\cite%
	\renewcommand{\cite}[1]{\normalfont\fontfamily{ptm}\selectfont\tiny\citeold{##1}}%
}
\newcolumntype{Y}{>{\centering\arraybackslash}X}

\titlecontents{section}
[1.5em]
{\addvspace{-0.5pt}}
{\bfseries\contentslabel{2.3em}}
{\hspace*{-2.3em}\bfseries}
{\bfseries\titlerule*[.5pc]{.}\contentspage}
\titlecontents{subsection}
[3.8em]
{\addvspace{-2.2pt}}
{\contentslabel{2.3em}}
{\hspace*{-2.3em}}
{\titlerule*[.5pc]{.}\contentspage}

\newtcolorbox{planbox}[1]{
    colback=gray!5,
    colframe=gray!75,
    title=#1,
    fonttitle=\bfseries
}

\tcbset{colback=white}
\colorlet{titleblue}{blue!80!black}
\colorlet{titlered}{red!80!black}
\colorlet{titlegreen}{green!80!black}
\colorlet{darkgreen}{green!50!black}

\definecolor{logoRed}{HTML}{CB2F10}
\definecolor{logoBlue}{HTML}{1A4E8A}
\definecolor{logoCyan}{HTML}{56BBCC}

\definecolor{genLLM}{RGB}{250,250,250}
\definecolor{specLLM}{RGB}{230,230,230}
\definecolor{ourLLM}{RGB}{230,242,230}
\definecolor{spaLLM}{RGB}{255,249,196}

\definecolor{genVLM}{RGB}{255,255,255}
\definecolor{ourVLM}{RGB}{210,210,210}
\DeclareCaptionFont{times}{\fontfamily{ptm}\selectfont}

\tcbset{
  aibox/.style={
    width=\linewidth,
    top=10pt,
    colback=white,
    colframe=black,
    colbacktitle=black,
    enhanced,
    center,
    attach boxed title to top left={yshift=-0.1in,xshift=0.15in},
    boxed title style={boxrule=0pt,colframe=white,},
  }
}

\newtcolorbox{AIbox}[2][]{aibox,title=#2,#1}

\usepackage{tikz}
\makeatletter
\definecolor{applegreen}{rgb}{0.55, 0.71, 0.0}

\abstract{
While agentic AI and its core multimodal large language models (MLLMs) have demonstrated remarkable promise in language and visual reasoning across domains ranging from daily life to advanced scientific research, a profound gap remains between artificial and human intelligence. Despite the integration of powerful tools and advanced MLLMs, state-of-the-art AI agents frequently fail at foundational, seemingly simple tasks that a child can resolve with ease.
Inspired by the Wechsler Intelligence Scale for Children (WISC), we introduce \benchmarknamenc{}, the first psychometrically grounded interactive benchmark for evaluating cognitive age alignment in MLLM-based agents.
\benchmarknamenc{} systematically compares the reasoning performance of various MLLM-based interactive agents against age-specific human developmental stages, exposing where current agentic AI systems can and cannot simulate age-specific cognitive behavior.
}

\metadata[GitHub]{\url{https://github.com/PediaMedAI/ChildAgentEval}}
\metadata[Correspondence to]{xucao@pediamed.ai}

\definecolor{lightgray}{rgb}{0.95, 0.95, 0.95}

\definecolor{baselinecolor}{gray}{.9}

\begin{document}

\maketitle

\section{Introduction}

Multimodal Large Language Model (MLLM) agents are increasingly integrated into social and educational environments to interact with users at distinct developmental stages~\citep{Zhang2023.07.10.23292373, singhal2023large, luo2025largelanguagemodelagent, boiko2023emergent, chen2025mlrbench}, particularly children and adolescents~\citep{kasneci2023chatgpt, piaget1952origins}. While the prevailing paradigm for AI development emphasizes maximizing task performance by leveraging sophisticated reasoning and vast knowledge, this approach is often counterproductive in child-centered contexts \citep{Park2023Generative}. For a child-facing tutor, technical correctness is only a baseline; true effectiveness depends on developmental alignment~\citep{Kail1991Developmental, cowan2010magical, mcgrew2009chc}. An agent that consistently employs adult-level abstractions or complex reasoning chains may fail to scaffold learning within a child's Zone of Proximal Development~\citep{vygotsky1978mind, lyons1984defining}. Such a system often provides explanations that transcend the developmental limits of the user's cognitive grasp~\citep{piaget1952origins}, missing the opportunity to address the child's specific confusion. This necessitates a shift from merely optimizing accuracy to behavioral calibration, posing the question of whether an AI agent can intentionally align its reasoning complexity, memory retention, and communicative style with a target developmental age.

This question is especially important for pediatric and adolescent users due to the high variability in user memory, attention, and reasoning. 
Middle childhood and early adolescence are critical periods for cognitive development and identity formation~\citep{eccles1999development}, and recent child-facing AI systems increasingly target tutoring, safety, childcare, and developmental interaction scenarios~\citep{murali2026evaluatingllmsafetychild,nayeem2024kidlmadvancinglanguagemodels,liu2025benchmarkingllmsmimickingchildcaregiver}. 
In such settings, technical correctness alone can be misleading. 
Agents relying on adult-level abstraction frequently exceed a child's cognitive limits and provide mismatched guidance. For child-facing AI, the primary objective shifts from raw problem-solving power to cognitive simulation: the ability to align its communicative and reasoning style with the developmental state of its partner.

Current evaluation paradigms provide limited tools for answering this question. 
Most agent benchmarks measure whether models solve tasks correctly, treating higher accuracy and more advanced task completion as uniformly better~\citep{phan2025humanity,lu2022learn}. 
Even evaluations of educational, healthcare, child-facing, and interactive AI systems rarely ask whether the model's reasoning process is developmentally appropriate for a specific user~\citep{kasneci2023chatgpt,Zhang2023.07.10.23292373,singhal2023large,murali2026evaluatingllmsafetychild,nayeem2024kidlmadvancinglanguagemodels,liu2025benchmarkingllmsmimickingchildcaregiver}. 
Consequently, an agent may appear highly capable yet remain poorly calibrated for children by using advanced vocabulary, adult-level abstractions, excessive information retention, or developmentally inconsistent strategies. While standard age prompting is a common shortcut, it remains unclear if asking a model to "act like a child" alters its underlying cognitive behavior or merely its surface style.

We study this problem through the lens of \textbf{\textit{cognitive age alignment}}, the ability of an interactive agent to produce behavior matched t
o a target stage of human cognitive development. Developmental alignment is not uniform capability reduction. Rather than degrading performance across all tasks, an aligned agent applies structured cognitive constraints: younger targets exhibit simpler language, restricted working memory, and specific error patterns, whereas older targets demonstrate progressively stronger reasoning and complex explanations \citep{piaget1952origins, cowan2010magical,Gathercole1999CognitiveAT}. This requires evaluating not only aggregate accuracy, but also whether performance, language, memory, and reasoning profiles change systematically with age.\looseness-1

To enable this evaluation,  we introduce \benchmarknamenc{}, an interactive benchmark for measuring developmental alignment in MLLM-based agents, inspired by the Wechsler Intelligence Scale for Children (WISC-IV) \citep{wechsler2003wisc}.  Instead of reproducing protected clinical items, ChildAgentEval draws on the WISC-IV framework, ensuring that its web-based tasks are informed by the target cognitive constructs that cover verbal comprehension, perceptual and fluid reasoning, and working memory.  
Rather than evaluating models only through final-answer accuracy, \benchmarknamenc{} measures age-normed composite scores, subtest-level behavior, trajectory-level developmental trends, and language complexity across target age conditions. This design allows us to ask whether an agent's behavior becomes meaningfully age-ordered, or whether the model continues to operate at its default capability level regardless of the requested age.
We further propose a \textbf{skill-guided distillation strategy} that translates empirical developmental markers into executable cognitive constraints. Beyond role-play prompts, our method specifies age-appropriate limits on reasoning strategies, memory load, linguistic complexity, and task-solving behavior. These constraints act as cognitive filters that guide the agent toward behavior consistent with the target developmental band.  Experiments on multimodal agents demonstrate that while standard prompting yields flat trajectories, our distillation method facilitates robust age differentiation.

Our experiments reveal three main findings. First, standard age prompting does not reliably induce developmental alignment: most models continue to maximize correctness and produce weak or irregular age trajectories. Second, skill guidance improves developmental differentiation in stronger proprietary models, producing more monotonic score trajectories and more age-sensitive language patterns. Third, alignment remains uneven across cognitive domains. Language-mediated behavior is relatively easy to control, while working memory, perceptual reasoning, and processing-speed behaviors remain difficult to calibrate because current MLLM architectures lack human-like limits on memory, attention, and visual processing. Together, these findings show that developmental alignment requires more than asking agents to act younger; it requires explicit constraints on how agents perceive, remember, reason, and communicate.
Our contributions are as follows:
\begin{enumerate}[label=(\arabic*), leftmargin=*,itemsep=0.5ex, parsep=0pt, topsep=0pt]
\item We define \textbf{cognitive age alignment} as a novel challenge, shifting the evaluation focus from maximizing raw capability to calibrating agent behaviors against human developmental structures.
\item We build {\benchmarknamenc{}}, a WISC-inspired interactive evaluation framework for measuring whether MLLM-based agents can align with target developmental ages across psychometrically grounded cognitive domains.
\item We introduce a data-driven \textbf{skill-guided distillation} strategy that converts developmental markers into executable cognitive constraints on language, memory, reasoning, and task-solving behavior.
\item We empirically demonstrate that standard prompting fails to produce stable developmental trajectories, whereas our distillation strategy significantly improves age differentiation and reveals current LLM limitations in calibrating working memory and visuospatial reasoning.
\end{enumerate}

\section{Related Works}
\label{sec:relatedwork}
\subsection{Psychometric and Cognitive Evaluation of LLMs and MLLMs}

In recent years, an increasing number of studies focus on benchmarking LLMs and VLMs through psychological and cognitive assessments~\citep{cao2024visual,li2026toward}. This goes beyond traditional paradigms. For more general assessments with a broader scope, examples include the IQ EQ PQ evaluation framework, which is an evaluation framework based on human perspectives \citep{wang2025benchmarkllmsevaluationanthropomorphic}. Other works evaluate state-of-the-art VLMs using the Wisconsin Card Sorting Test (WCST), a classical measurement method for set shifting ability \citep{hao2025visuallargelanguagemodels}. Additionally, MLR Bench contains over 400 carefully curated tasks to achieve a comprehensive evaluation of the end-to-end research capabilities of agents \citep{chen2025mlrbench}. Other works, such as AgentBoard test 11 open source models by focusing on fine-grained action metrics rather than relying solely on accuracy and scores \citep{ma2024agentboardanalyticalevaluationboard}. IQBench proposes a vision-centric approach to evaluate the performance of VLMs in standardized visual intelligence tests \citep{pham2025iqbenchsmartvisionlanguagemodels}. 
At the same time, more studies investigate clinical cognitive tests for LLMs \citep{Zhang2023.07.10.23292373}. From the perspective of psychometrics, KidGym draws on the Wechsler Intelligence Scale to propose a benchmark containing 12 unique tasks. The abilities targeted by these tasks can evaluate and reflect the stages of child cognitive development \citep{ye2026childrensintelligencetestspose}.

Recent work has also begun to comprehensively compare generative models against population-normed benchmarks, such as estimating the normative intelligence of language models \citep{ilic2024estimating, galatzer2024cognitive} and systematically evaluating LLMs using human psychometric tests \citep{jung2026psychometric}. Further research demonstrates that psychometric comparison to human normative distributions is becoming a viable evaluation direction for foundation models~\citep{galatzer2024cognitive, king2023iq, wasilewski2024mmiq, huang2024wais}. However, that line of work focuses on adult-oriented cognitive benchmarks and does not examine developmental calibration in an interactive agent setting. Unlike these studies, our work is not merely an intelligence quotient benchmark. Instead, it features age stratification and grounding in developmental psychology within an agentic multi step setting. Furthermore, we apply skill distillation from real child interaction data and evaluate agents using both scores and error patterns.

\subsection{LLMs as Cognitive Models and Human Simulators}

Meanwhile, a large amount of research has begun to leverage large language models (LLMs) and generative agents as computational tools for simulating human cognition, ranging from general behavioral patterns to more abstract psychological processes \citep{xie2024largelanguagemodelagents,LI2025108687,https://doi.org/10.1111/cogs.70106}. Centaur fine tunes a computational model capable of predicting and simulating human behavior using the Psych 101 dataset \citep{binz2025centaurfoundationmodelhuman}. Other studies design a framework that uses LLMs as psychological simulators for role characters to simulate how these characters explore various scenarios or conduct cognitive modeling \citep{Lin_2026}. Similar frameworks design realistic senior executive agents using LLMs based on real communication content and moral foundations \citep{garzonvico2026usinglargelanguagemodels}. In addition, some works focus on whether the Generative Agent Based Model (GABM) can establish Theory of Mind (ToM) in the real world \citep{lombardi2025doingthingswordsrethinking}. These existing works mostly focus on adults and lean toward general behavioral or social simulation. They rarely address developmental cognition and lack psychometric calibration.

\subsection{Child-focused LLM Simulation and Safety}

Regarding child cognitive simulation, related works have started to evaluate the safety and language patterns of large language models. For instance, ChildSafe evaluates the safety of language models by simulating child agents in different developmental stages \citep{murali2026evaluatingllmsafetychild, jiao2025safechildllm,xing2025sproutbench} align models with the unique preferences of young users \citep{nayeem2024kidlmadvancinglanguagemodels, xing2025sproutbench,jiao2025safechildllm}. Furthermore, significant efforts have been directed at analyzing child-caregiver interactions, evaluating whether LLMs can replicate these linguistic features \citep{liu2025benchmarkingllmsmimickingchildcaregiver, jarvilehto2026largelanguagemodel} and automating the grammatical annotation of transcribed conversations \citep{nikolaus2024automaticannotationgrammaticalitychildcaregiver}. researchers have begun investigating interactive simulations and developmental cognition. This includes deploying AI-driven child avatars for dynamic interviewing tasks \citep{jarvilehto2026largelanguagemodel}, comparing LLM architectures to human cognitive development across age groups \citep{demetriou2025speciesofmind}, and adapting classical developmental psychology experiments to probe the computational capabilities of models like LaMDA and GPT \citep{kosoy2023comparingmachineschildrenusing, yiu2024transmission}. Currently, there is no systematic research on how to distill age specific skills from real child data and inject these mechanisms into an agent. Moreover, existing literature lacks approaches that use psychometric benchmarks to measure whether an agent truly reasons like a specific age group.

\section{ChildAgentEval}
\label{section:benchmark}

\begin{figure}[t!]
    \centering
    \includegraphics[width=\textwidth]{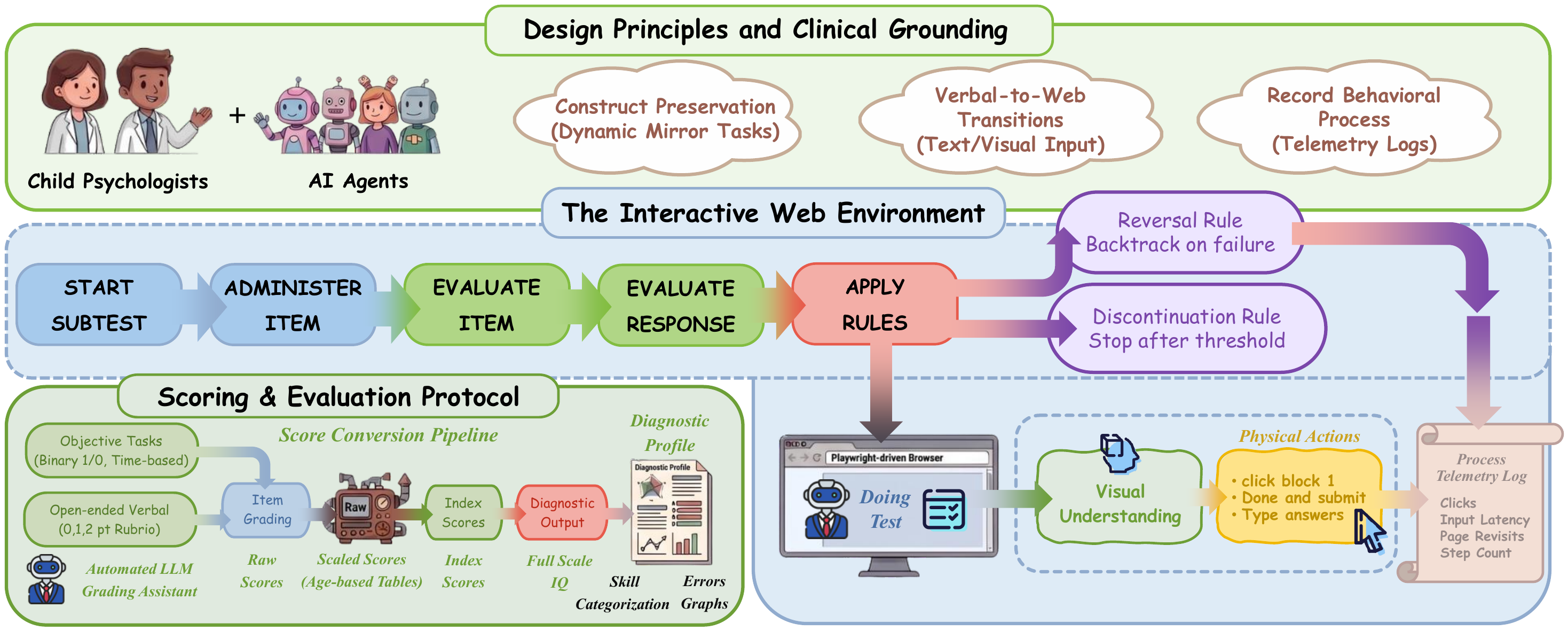}
    \caption{\textbf{The comprehensive architecture of \benchmarknamenc.} The framework adapts human-administered assessment design principles into an interactive web evaluation pipeline, integrating test administration, behavioral logging, and standardized scoring.}
    \label{fig:childagentbench_arch}
\end{figure}

While the WISC serves as the gold standard for pediatric intelligence assessment~\citep{wechsler2003wisc}, its format was originally designed for human clinical administration rather than AI-based evaluation. Accordingly, adapting Wechsler-inspired cognitive constructs for agent-based evaluation is critical. We therefore develop web-based tasks conceptually aligned with standard cognitive assessments, in which AI agents must execute interactive browser actions, maintain working memory, and make sequential decisions (Fig. \ref{fig:childagentbench_arch} for an overview).

\begin{figure}[!t] 
    \centering
    \includegraphics[width=\textwidth]{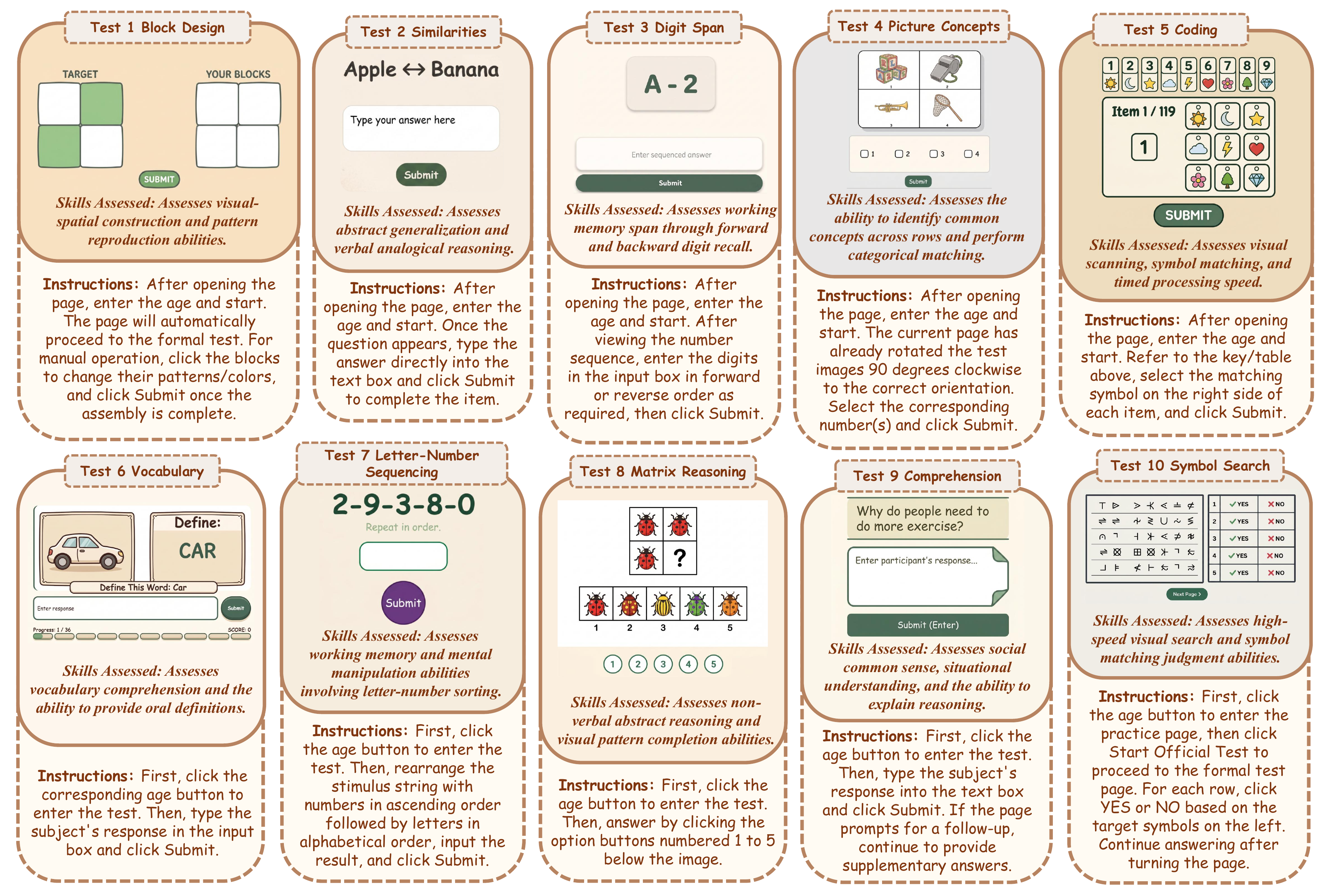}
    \caption{\textbf{Overview of the ten interactive subtests in ChildAgentEval.} 
    Each panel illustrates the dynamic web interface, the specific cognitive skills assessed, and the required physical interactions that the AI agent must perform to simulate human cognitive problem-solving.}
    \label{fig:subtests_overview}
\end{figure}

\paragraph{Design Principles and Grounding.}
The platform consists of ten interactive subtests mapped to the Cattell-Horn-Carroll (CHC) intelligence model \citep{mcgrew2009chc}, evaluating verbal abstraction, vocabulary, comprehension, fluid and visual reasoning, working memory, and processing speed. Specifically, crystallized intelligence (Gc) includes Similarities (Test 2), Vocabulary (Test 6), and Comprehension (Test 9). The fluid reasoning and visual-spatial dimension (Gf/Gv) addresses rule induction and spatial problem solving via Block Design (Test 1), Picture Concepts (Test 4), and Matrix Reasoning (Test 8). Working memory (WM) involves information retention and manipulation through Digit Span (Test 3) and Letter-Number Sequencing (Test 7), while processing speed (PSI) measures execution through Coding (Test 5) and Symbol Search (Test 10). Figure~\ref{fig:subtests_overview} provides a visual overview of these subtests and their interactive formats. To ensure validity, the platform was developed in collaboration with child psychologists, who reviewed the task design, age stratification, and scoring procedures to support developmentally appropriate assessment standards.

Adapting clinical scales to web environments involves three structural principles. First, construct preservation maps cognitive abilities into dynamic interactive tests instead of static items~\citep{sainz2023contamination} ; for example, the coding test uses a dynamic symbol table with strict time limits. Second, we operationalize verbal administration as web interactions by using text inputs and presenting sequences across separate pages to prevent context window leakage~\citep{gong2024wm,hu2025memory}. For spatial tasks such as Block Design, numbered Document Object Model (DOM) labels convert physical clicks into numerical selections, ensuring that errors reflect reasoning deficits rather than visual localization failures. Third, the system records the complete behavioral process by logging granular data like clicks, latency, and step counts. These telemetry logs provide process-level insights into rule retention or visual distraction. Finally, the platform is restricted to secure research settings.\looseness-1

\textbf{The Interactive Web Environment.}
Built upon a Finite State Machine architecture, the system operates each subtest independently to execute the standard administration protocol. This includes the Reversal rule, which reverts the agent to foundational items if it fails the first two questions at a higher age starting point, and the Discontinuation rule ends a subtest once a predefined number of consecutive zero scores is reached.
The testing environment utilizes Playwright to drive a simulated browser, requiring the agent to rely on visual understanding and physical actions such as clicking, typing, and selecting. Throughout this process, the system automatically logs interaction metrics and state transition graphs to strictly record the behavior of the agent.

\textbf{Evaluation Protocol.}
\label{sec:cognitive_domain}
The platform evaluates four primary cognitive factors. Gc, Gf/Gv, WM and PSI.
To ensure the evaluation follows a grounded developmental trajectory spanning 6--16 years, the system enforces age-specific start items and difficulty levels according to clinical guidelines. By encoding cognitive constraints derived from empirical data, ChildAgentEval provides a holistic framework to pinpoint exactly where the reasoning capabilities of an agent align with human cognitive development.
The scoring protocol evaluates items based on their specific task formats. Objective subtests (Picture Concepts, Matrix Reasoning, Block Design, Symbol Search) and early vocabulary items apply a strict binary scoring mechanism, awarding one point for a correct action or exact keyword match. For processing speed tests (Coding), the score is the total number of correct operations executed within the time constraint. For open-ended verbal reasoning tests (advanced Vocabulary, Similarities, Comprehension), responses are graded against a standard zero, one, or two-point rubric. We use GPT-5.4 as a grading assistant for processing linguistic outputs at scale, but all automated scores for open-ended questions undergo mandatory verification by independent human raters.

Following the item-level grading, raw scores from each subtest are mapped to scaled scores using established age-based normative tables. These scaled scores are aggregated to compute the respective Index Scores for the four primary cognitive domains, which are then synthesized into the Full Scale Intelligence Quotient (FSIQ)~\citep{klein2024verbal,galatzer2024cognitive}. By implementing this standard conversion procedure, the system ensures the measurement is statistically grounded. The final benchmark output reports these detailed performance metrics alongside systematically categorized error tags.\looseness-1


\section{Age-Specific Cognitive Skill Distillation}
\label{section: exploration}
The age-specific settings used in ChildAgentEval do not rely on subjective construction based on current stereotypes of children or teenagers, or directly designing simple system role prompts. Instead, we extract age-specific cognitive skills from real interaction data of children and adolescents. We construct a parameterized cognitive distillation architecture that translates human cognitive development features into executable constraints for large language model agents.

\noindent \textbf{Data Collection and Age Slicing Normalization.}

To accurately capture the cognitive features of different developmental stages, we integrate a multi-source corpus covering ages 6 to 17. Detailed information regarding the specific datasets and data splits is provided in the Appendix \ref{app:data_details}. For lower age groups, we rely on spoken and multimodal interaction data to capture daily vocabulary boundaries, immediate attention spans, and self-repair markers. For higher age groups, we use classroom discussions, psychological interviews, and narrative writing texts to capture abstract vocabulary use, long-range logical reasoning, and adolescent egocentric bias. During data processing, we strictly filter the dialogue corpora to retain only the original utterances of minors, eliminating cognitive contamination from adult guidance. Finally, we apply uniform normalization to all texts to calculate basic linguistic metrics and balance the data distribution across test types.

\noindent \textbf{Cognitive Profile Vector Representation.}

We model the features of each age group as a cognitive profile vector rather than making the model imitate a speaking tone. This vector contains six core dimensions \citep{mcgrew2009chc, jarvilehto2026largelanguagemodel}. As introduced in \S~\ref{sec:cognitive_domain}, five of these dimensions are Gc, Gf, Gv, WM, and PSI. We use these to parameterize the upper limit of vocabulary abstraction, the depth of logical reasoning, the capacity for temporary information retention, the degree of reliance on visual representation, and the speed and attentional stability of cognitive processing. We also retain a Social dimension as an auxiliary control variable to capture perspective switching and the degree of egocentric bias in open-ended social reasoning. This structure allows us to design the upper limit of cognitive ability, typical strategy paths, and error patterns that are prone to occur in a specific age group.

\noindent \textbf{Two Stage Skill Distillation Pipeline.}
\begin{figure}[t]
    \centering
    \includegraphics[width=\linewidth]{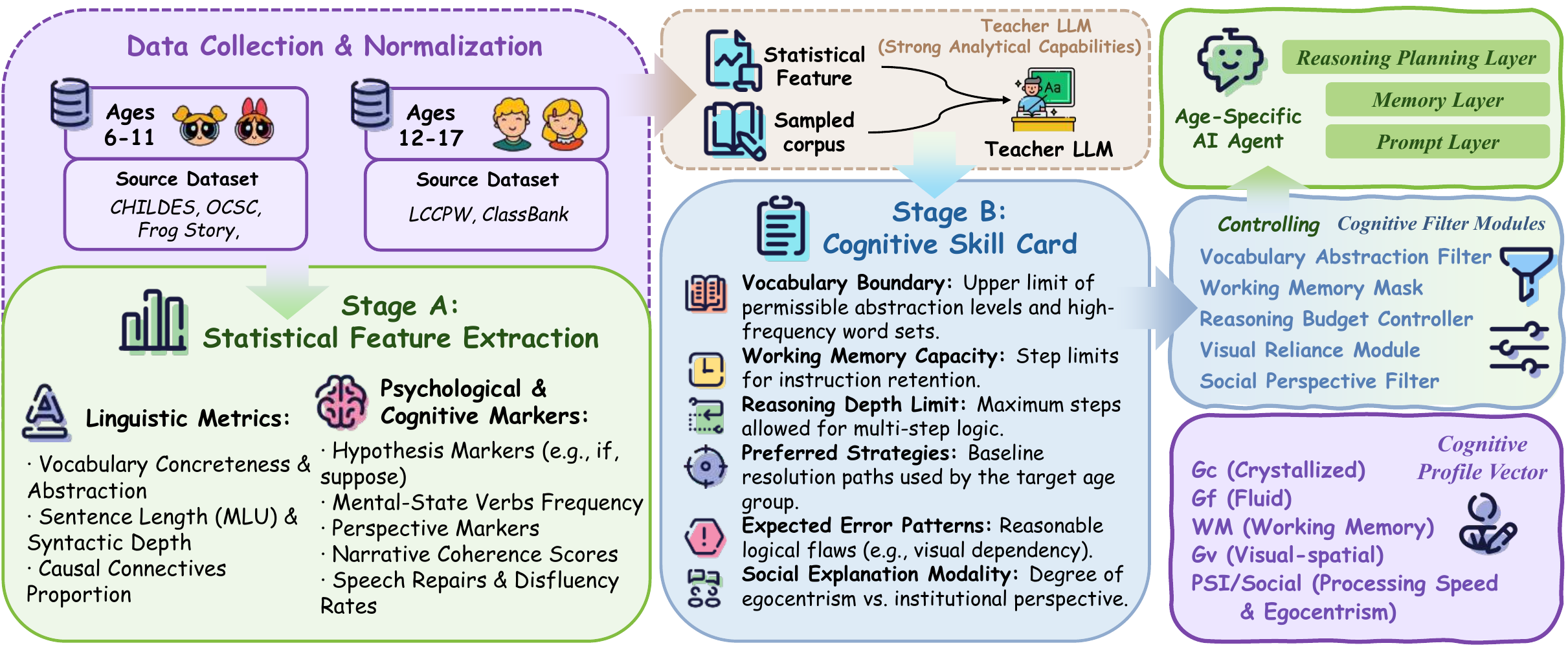} 
    \caption{\textbf{The Two-Stage Skill Distillation Pipeline.} Stage A involves the statistical extraction of linguistic metrics and psychological markers from raw corpora. Stage B uses a Teacher LLM to distill these statistical features into structured Cognitive Skill Cards.}
    \label{fig:skill_pipeline}
\end{figure}

To transform the original corpora into these cognitive profile vectors, we design a two-stage distillation pipeline as shown in Fig. \ref{fig:skill_pipeline}. The first stage focuses on statistical feature extraction using a combination of transcript-specific analyzers and semantic natural language processing toolkits. We measure lexical diversity, semantic concreteness \citep{brysbaert2014concreteness}, sentence length, grammatical depth, and expression fluency. We also apply custom lexical matching to quantify the frequencies of mental state verbs, causal connectives, and conditional clauses. In the second stage, we input these statistical distribution results and sampled children corpus fragments into a teacher language model. Through strict instruction constraints, the teacher model outputs standardized cognitive skill cards that specify the high-frequency vocabulary boundaries of the target age group, the upper depth limit of multi-step reasoning, preferred resolution strategies, and expected logical error patterns.

\noindent \textbf{Cognitive Filter Module and Agent Integration.}

To implement the distilled cognitive skills, we design five cognitive filter modules and inject them into the prompt layer, memory layer, and reasoning planning layer of the agent. The vocabulary abstraction filter controls syntactic complexity and limits lower age agents from using academic concepts. The working memory mask physically simulates shorter memory spans by restricting retained information across pages or injecting memory noise. The reasoning budget controller intervenes in the chain of thought, restricting lower age agents to direct observation matching while allowing higher age agents to execute hypothesis verification. The visual reliance module reproduces cognitive biases, making lower age agents easily misled by physical arrangement illusions such as height and area. The social perspective filter restricts the standpoint of the agent when explaining social norms, such as limiting young children to first-person explanations while allowing adolescents to use institutional perspectives. The system automatically loads the corresponding skill configuration based on the target age. To ensure psychometric validity, we finetune the intervention strength of each module on an independent calibration set to approximate human norms, and we evaluate generalization on a held-out test set. \looseness-1
\section{Experiments}
Our experiments evaluate the capacity of MLLM-based interactive agents to align with human developmental trajectories under standardized psychometric conditions. We address two questions: (1) whether data-grounded skill distillation induces age-appropriate reasoning and behavior more effectively than standard prompting \citep{Park2023Generative} ; and (2) if these alignment patterns and cognitive deficits are consistent across diverse proprietary and open-weight architectures.\looseness-1

\textbf{Implementation Details and Backbone Models}:  The assessment measures performance across four specific anchor ages: 7, 10, 13, and 16 years old. Detailed procedures regarding the interactive administration logic and the overall environment design are provided in \S~\ref{section:benchmark}. The experiments compare two settings: a Baseline condition utilizing standard prompting with age labels, and a Skill-Guided condition that applies distilled age-specific skill configurations detailed in \S~\ref{section: exploration}. We evaluate interactive agents instantiated from both proprietary and open-weight backbone models. The proprietary backbones are GPT-5.4, Gemini-3.1-Pro, Gemini-3.1-Flash-Lite, and Qwen-3.6-Plus. The open-weight backbones are Qwen-3.5-27B and Gemma-4-31B. For fair comparison and reproducibility, we utilize greedy decoding across all model API calls by setting the temperature to 0.0. \looseness-1
Each run generates structured result files and detailed action logs for every age, permitting the analysis of both final scores and intermediate behaviors. We provide more implementation details on Appendix \ref{app:implement}

\textbf{Evaluation Metrics}: We report four groups of evaluation metrics. First, we normalize raw total and subtest scores by their theoretical maximums. Second, to measure developmental differentiation, we evaluate total-score trajectory monotonicity and associated trend statistics. Third, we aggregate subtests into Gc, WM, Gf/Gv, and PSI factor scores, computing their age-normed deviation z-scores alongside FSIQ. Fourth, we assess linguistic age fidelity in open-ended responses using mean utterance length, lexical diversity, and causal or definitional constructions \citep{brown1973firstlanguage}. This establishes a dual-layered evaluation: normalized scores capture absolute within-evaluation task performance, while age-normed z-scores reveal the agent's composite deviation from human developmental norms.\looseness-1

\begin{table*}[t!]
\centering
\caption{\textbf{Normalized Benchmark Subtest Scores and Age-Normed WISC Composite $z$-Scores for Proprietary Models.}
Total and T1--T10 are normalized benchmark scores; composite columns report age-normed deviations.}
\label{tab:joint_norm_sd_report_prop}

\setlength{\tabcolsep}{3.2pt}
\renewcommand{\arraystretch}{1.08}
\scriptsize

\resizebox{0.97\textwidth}{!}{%
\begin{tabular}{llcc cccc ccc cccc ccc}
\toprule
\multirow{2}{*}{\textbf{Method}} &
\multirow{2}{*}{\textbf{Setting}} &
\multicolumn{2}{c}{\textbf{Overall}} &
\multicolumn{4}{c}{\textbf{Gc}} &
\multicolumn{3}{c}{\textbf{WM}} &
\multicolumn{4}{c}{\textbf{Gf/Gv}} &
\multicolumn{3}{c}{\textbf{PSI}} \\
\cmidrule(lr){3-4}
\cmidrule(lr){5-8}
\cmidrule(lr){9-11}
\cmidrule(lr){12-15}
\cmidrule(lr){16-18}
& & \textbf{Total} & \textbf{FSIQ z}
& \textbf{Gc z} & T2 & T6 & T9
& \textbf{WM z} & T3 & T7
& \textbf{Gf/Gv z} & T1 & T4 & T8
& \textbf{PSI z} & T5 & T10 \\
\midrule

\rowcolor{gray!8}
\multicolumn{18}{l}{\textbf{GPT-5.4}} \\
& Baseline (7) & \cellcolor[HTML]{DDEFD8}0.53 & \textbf{2.07} & \textbf{5.33} & 0.89 & \cellcolor[HTML]{DDEFD8}0.93 & \cellcolor[HTML]{DDEFD8}1.00 & \textbf{5.33} & \cellcolor[HTML]{DDEFD8}1.00 & \cellcolor[HTML]{DDEFD8}1.00 & \textbf{-2.00} & 0.09 & 0.07 & 0.40 & \textbf{-1.40} & \cellcolor[HTML]{DDEFD8}0.29 & 0.22 \\
& Baseline (10) & 0.49 & \textbf{1.33} & \textbf{4.93} & 0.89 & 0.87 & 0.90 & \textbf{5.33} & \cellcolor[HTML]{DDEFD8}1.00 & \cellcolor[HTML]{DDEFD8}1.00 & \textbf{-2.93} & 0.09 & 0.11 & 0.34 & \textbf{-2.33} & 0.20 & 0.20 \\
& Baseline (13) & 0.46 & \textbf{0.13} & \textbf{2.93} & 0.89 & 0.71 & 0.81 & \textbf{4.73} & \cellcolor[HTML]{DDEFD8}1.00 & \cellcolor[HTML]{DDEFD8}1.00 & \textbf{-3.20} & \cellcolor[HTML]{DDEFD8}0.15 & 0.11 & 0.29 & \textbf{-3.20} & 0.21 & 0.17 \\
& Baseline (16) & 0.52 & \textbf{0.20} & \textbf{2.47} & 0.91 & 0.78 & 0.81 & \textbf{4.47} & \cellcolor[HTML]{DDEFD8}1.00 & \cellcolor[HTML]{DDEFD8}1.00 & \textbf{-3.20} & \cellcolor[HTML]{DDEFD8}0.15 & 0.21 & 0.26 & \textbf{-2.13} & 0.20 & \cellcolor[HTML]{DDEFD8}0.57 \\
\cmidrule(lr){2-18}
& Skill-Guided (6--8) & 0.41 & \textbf{1.40} & \textbf{3.87} & 0.68 & 0.47 & 0.88 & \textbf{5.33} & \cellcolor[HTML]{DDEFD8}1.00 & 0.97 & \textbf{-2.13} & 0.06 & 0.14 & 0.31 & \textbf{-1.93} & 0.22 & 0.15 \\
& Skill-Guided (9--11) & 0.42 & \textbf{0.40} & \textbf{2.80} & 0.77 & 0.49 & 0.90 & \textbf{5.33} & \cellcolor[HTML]{DDEFD8}1.00 & \cellcolor[HTML]{DDEFD8}1.00 & \textbf{-2.80} & 0.09 & 0.21 & 0.29 & \textbf{-2.73} & 0.18 & 0.15 \\
& Skill-Guided (12--14) & 0.49 & \textbf{0.33} & \textbf{2.93} & 0.82 & 0.78 & 0.81 & \textbf{4.73} & \cellcolor[HTML]{DDEFD8}1.00 & \cellcolor[HTML]{DDEFD8}1.00 & \textbf{-2.93} & 0.09 & 0.32 & 0.31 & \textbf{-2.53} & 0.19 & 0.33 \\
& Skill-Guided (15--17) & 0.50 & \textbf{0.07} & \textbf{2.60} & 0.95 & 0.81 & 0.76 & \textbf{4.47} & \cellcolor[HTML]{DDEFD8}1.00 & \cellcolor[HTML]{DDEFD8}1.00 & \textbf{-3.20} & 0.09 & 0.21 & 0.37 & \textbf{-3.20} & 0.20 & 0.30 \\

\midrule
\rowcolor{gray!8}
\multicolumn{18}{l}{\textbf{Gemini-3.1-Pro}} \\
& Baseline (7) & 0.37 & \textbf{1.07} & \textbf{0.27} & 0.91 & 0.12 & 0.12 & \textbf{5.33} & \cellcolor[HTML]{DDEFD8}1.00 & \cellcolor[HTML]{DDEFD8}1.00 & \textbf{0.53} & 0.09 & 0.82 & 0.40 & \textbf{-1.93} & 0.11 & 0.30 \\
& Baseline (10) & 0.48 & \textbf{1.87} & \textbf{4.47} & 0.86 & 0.79 & 0.88 & \textbf{5.33} & \cellcolor[HTML]{DDEFD8}1.00 & \cellcolor[HTML]{DDEFD8}1.00 & \textbf{-0.73} & \cellcolor[HTML]{DDEFD8}0.15 & 0.79 & 0.26 & \textbf{-2.93} & \cellcolor[HTML]{F6D9D9}0.02 & 0.22 \\
& Baseline (13) & 0.44 & \textbf{0.27} & \textbf{2.20} & 0.64 & 0.75 & 0.86 & \textbf{4.73} & \cellcolor[HTML]{DDEFD8}1.00 & \cellcolor[HTML]{DDEFD8}1.00 & \textbf{-1.87} & \cellcolor[HTML]{DDEFD8}0.15 & 0.68 & 0.26 & \textbf{-3.67} & 0.05 & 0.15 \\
& Baseline (16) & 0.39 & \textbf{-0.33} & \textbf{0.40} & 0.86 & 0.24 & 0.81 & \textbf{4.47} & \cellcolor[HTML]{DDEFD8}1.00 & \cellcolor[HTML]{DDEFD8}1.00 & \textbf{-1.87} & 0.09 & 0.79 & 0.29 & \textbf{-3.67} & \cellcolor[HTML]{F6D9D9}0.02 & 0.15 \\
\cmidrule(lr){2-18}
& Skill-Guided (6--8) & 0.32 & \textbf{0.47} & \textbf{1.93} & 0.66 & 0.62 & 0.07 & \textbf{1.47} & 0.47 & 0.67 & \textbf{0.40} & 0.09 & 0.82 & 0.37 & \textbf{-2.73} & 0.08 & 0.15 \\
& Skill-Guided (9--11) & 0.38 & \textbf{0.07} & \textbf{1.47} & 0.84 & 0.75 & 0.10 & \textbf{1.80} & 0.58 & 0.83 & \textbf{-0.73} & 0.09 & 0.79 & 0.37 & \textbf{-2.93} & 0.08 & 0.22 \\
& Skill-Guided (12--14) & 0.45 & \textbf{0.60} & \textbf{3.27} & 0.91 & 0.74 & 0.81 & \textbf{2.13} & 0.63 & \cellcolor[HTML]{DDEFD8}1.00 & \textbf{-0.87} & 0.09 & \cellcolor[HTML]{DDEFD8}0.89 & 0.26 & \textbf{-3.20} & 0.08 & 0.22 \\
& Skill-Guided (15--17) & 0.45 & \textbf{0.13} & \textbf{1.47} & 0.73 & 0.79 & 0.81 & \textbf{4.47} & \cellcolor[HTML]{DDEFD8}1.00 & \cellcolor[HTML]{DDEFD8}1.00 & \textbf{-1.60} & 0.09 & 0.86 & 0.20 & \textbf{-3.67} & 0.03 & 0.15 \\

\midrule
\rowcolor{gray!8}
\multicolumn{18}{l}{\textbf{Gemini-3.1-Flash-Lite}} \\
& Baseline (7) & 0.46 & \textbf{1.60} & \textbf{5.33} & 0.93 & 0.81 & \cellcolor[HTML]{DDEFD8}1.00 & \textbf{5.33} & \cellcolor[HTML]{DDEFD8}1.00 & \cellcolor[HTML]{DDEFD8}1.00 & \textbf{-2.13} & \cellcolor[HTML]{F6D9D9}0.03 & 0.21 & 0.29 & \textbf{-2.93} & 0.18 & \cellcolor[HTML]{F6D9D9}0.00 \\
& Baseline (10) & 0.47 & \textbf{1.07} & \textbf{4.47} & 0.93 & 0.74 & 0.90 & \textbf{5.33} & \cellcolor[HTML]{DDEFD8}1.00 & \cellcolor[HTML]{DDEFD8}1.00 & \textbf{-2.80} & 0.12 & 0.18 & 0.29 & \textbf{-2.73} & 0.16 & 0.17 \\
& Baseline (13) & 0.46 & \textbf{0.13} & \textbf{3.27} & 0.91 & 0.75 & 0.81 & \textbf{4.73} & \cellcolor[HTML]{DDEFD8}1.00 & \cellcolor[HTML]{DDEFD8}1.00 & \textbf{-3.47} & 0.06 & 0.21 & \cellcolor[HTML]{F6D9D9}0.14 & \textbf{-3.20} & 0.23 & 0.13 \\
& Baseline (16) & 0.44 & \textbf{-0.47} & \textbf{1.73} & 0.93 & 0.63 & 0.81 & \textbf{4.47} & \cellcolor[HTML]{DDEFD8}1.00 & \cellcolor[HTML]{DDEFD8}1.00 & \textbf{-3.60} & 0.12 & \cellcolor[HTML]{F6D9D9}0.04 & 0.17 & \textbf{-3.40} & 0.23 & 0.08 \\
\cmidrule(lr){2-18}
& Skill-Guided (6--8) & 0.36 & \textbf{0.67} & \textbf{3.87} & 0.41 & 0.71 & 0.86 & \textbf{5.33} & \cellcolor[HTML]{DDEFD8}1.00 & \cellcolor[HTML]{DDEFD8}1.00 & \textbf{-2.80} & \cellcolor[HTML]{F6D9D9}0.03 & 0.11 & \cellcolor[HTML]{F6D9D9}0.14 & \textbf{-3.40} & 0.08 & \cellcolor[HTML]{F6D9D9}0.00 \\
& Skill-Guided (9--11) & 0.45 & \textbf{0.87} & \textbf{4.47} & 0.77 & 0.88 & 0.90 & \textbf{5.33} & \cellcolor[HTML]{DDEFD8}1.00 & \cellcolor[HTML]{DDEFD8}1.00 & \textbf{-3.20} & 0.12 & 0.11 & \cellcolor[HTML]{F6D9D9}0.14 & \textbf{-2.93} & 0.15 & 0.10 \\
& Skill-Guided (12--14) & 0.47 & \textbf{0.33} & \textbf{4.20} & 0.95 & 0.90 & 0.81 & \textbf{4.73} & \cellcolor[HTML]{DDEFD8}1.00 & \cellcolor[HTML]{DDEFD8}1.00 & \textbf{-3.47} & 0.06 & 0.11 & 0.26 & \textbf{-3.40} & 0.20 & 0.07 \\
& Skill-Guided (15--17) & 0.48 & \textbf{0.00} & \textbf{3.07} & 0.93 & 0.88 & 0.81 & \textbf{4.20} & \cellcolor[HTML]{DDEFD8}1.00 & 0.97 & \textbf{-3.47} & 0.12 & 0.11 & 0.29 & \textbf{-3.40} & 0.19 & 0.20 \\

\midrule
\rowcolor{gray!8}
\multicolumn{18}{l}{\textbf{Qwen3.6-Plus}} \\
& Baseline (7) & 0.31 & \textbf{0.13} & \textbf{2.07} & 0.91 & \cellcolor[HTML]{F6D9D9}0.03 & 0.81 & \textbf{3.67} & 0.68 & 0.97 & \textbf{-1.73} & 0.06 & 0.14 & 0.49 & \textbf{-3.40} & 0.03 & 0.07 \\
& Baseline (10) & 0.31 & \textbf{-0.60} & \textbf{0.40} & 0.91 & 0.10 & 0.50 & \textbf{3.93} & 0.84 & 0.93 & \textbf{-2.27} & \cellcolor[HTML]{DDEFD8}0.15 & 0.18 & \cellcolor[HTML]{DDEFD8}0.51 & \textbf{-3.67} & 0.04 & 0.02 \\
& Baseline (13) & 0.28 & \textbf{-1.60} & \textbf{-0.93} & 0.95 & 0.16 & \cellcolor[HTML]{F6D9D9}0.02 & \textbf{1.93} & 0.63 & 0.93 & \textbf{-3.20} & \cellcolor[HTML]{F6D9D9}0.03 & 0.07 & 0.43 & \textbf{-2.93} & 0.06 & 0.27 \\
& Baseline (16) & 0.31 & \textbf{-1.60} & \textbf{-0.27} & \cellcolor[HTML]{DDEFD8}1.00 & 0.13 & 0.31 & \textbf{1.93} & 0.79 & 0.93 & \textbf{-3.33} & 0.09 & 0.11 & 0.40 & \textbf{-3.40} & 0.06 & 0.22 \\
\cmidrule(lr){2-18}
& Skill-Guided (6--8) & \cellcolor[HTML]{F6D9D9}0.25 & \textbf{-0.53} & \textbf{2.80} & 0.73 & 0.19 & 0.67 & \textbf{-1.00} & \cellcolor[HTML]{F6D9D9}0.32 & 0.37 & \textbf{-1.00} & 0.06 & 0.39 & 0.49 & \textbf{-3.67} & 0.03 & 0.03 \\
& Skill-Guided (9--11) & \cellcolor[HTML]{F6D9D9}0.25 & \textbf{-1.73} & \textbf{0.73} & \cellcolor[HTML]{F6D9D9}0.32 & 0.34 & 0.81 & \textbf{-2.20} & 0.37 & \cellcolor[HTML]{F6D9D9}0.20 & \textbf{-2.00} & 0.09 & 0.36 & \cellcolor[HTML]{DDEFD8}0.51 & \textbf{-3.67} & 0.04 & 0.05 \\
& Skill-Guided (12--14) & 0.33 & \textbf{-1.53} & \textbf{0.87} & 0.93 & 0.46 & 0.76 & \textbf{-0.60} & 0.53 & 0.67 & \textbf{-2.53} & 0.09 & 0.43 & 0.37 & \textbf{-3.67} & 0.03 & \cellcolor[HTML]{F6D9D9}0.00 \\
& Skill-Guided (15--17) & 0.37 & \textbf{-1.27} & \textbf{0.20} & 0.59 & 0.65 & 0.79 & \textbf{0.80} & 0.55 & 0.87 & \textbf{-1.87} & 0.09 & 0.64 & \cellcolor[HTML]{DDEFD8}0.51 & \textbf{-3.67} & 0.03 & 0.03 \\
\bottomrule
\end{tabular}%
}

\par\vspace{2pt}
\begin{minipage}{0.98\textwidth}
\footnotesize
\textit{Note.}
T denotes test index. `Total` and `T1--T10` report normalized benchmark scores derived directly from the web-based task outcomes. By contrast, `FSIQ z`, `Gc z`, `WM z`, `Gf/Gv z`, and `PSI z` report age-referenced deviation scores computed from WISC-style composite scores after age-based norm conversion. Specifically, raw subtest scores are first mapped to age-referenced scaled scores, these scaled scores are then aggregated into composite scores, and the reported deviation values are finally computed as $z=(S-100)/15$, where $S$ denotes the corresponding composite score on the WISC-style normative scale. For the skill-guided condition, the four rows correspond to target age bands 6--8, 9--11, 12--14, and 15--17. Soft green marks the highest value and soft red marks the lowest value among non-$z$ numeric columns within each table block.
\end{minipage}
\end{table*}

\subsection{Main Results and Age Trajectories}

\begin{figure*}[t]
\centering
\begin{minipage}{0.99\textwidth}
\centering
\captionsetup[subfigure]{labelfont={times}}
\begin{minipage}[t]{0.24\textwidth}
   
    \centering
    \includegraphics[width=\linewidth]{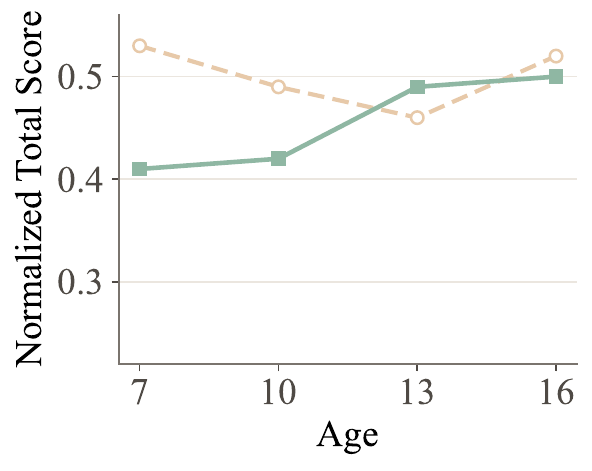}
    
    \subcaption{GPT-5.4}
\end{minipage}
\hfill
\begin{minipage}[t]{0.24\textwidth}
    \centering
    \includegraphics[width=\linewidth]{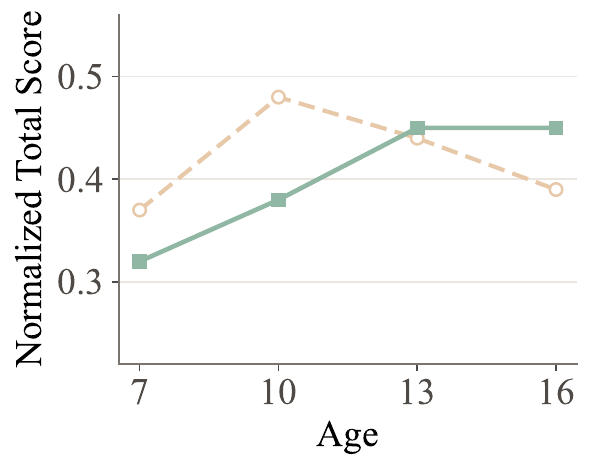}
    \subcaption{Gemini-3.1-Pro}
\end{minipage}
\hfill
\begin{minipage}[t]{0.24\textwidth}
    \centering
    \includegraphics[width=\linewidth]{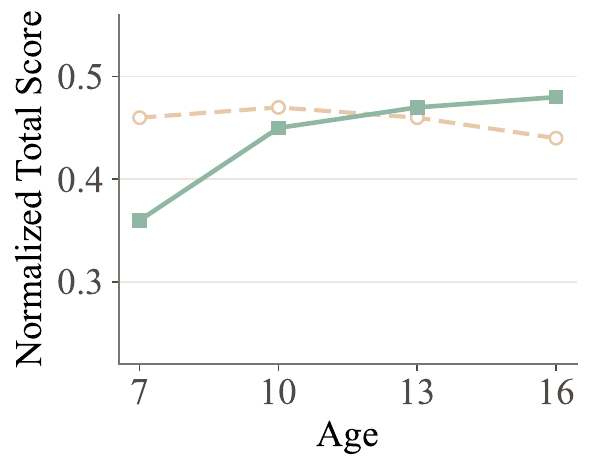}
    \subcaption{Gemini-3.1-Flash-Lite}
\end{minipage}
\hfill
\begin{minipage}[t]{0.24\textwidth}
    \centering
    \includegraphics[width=\linewidth]{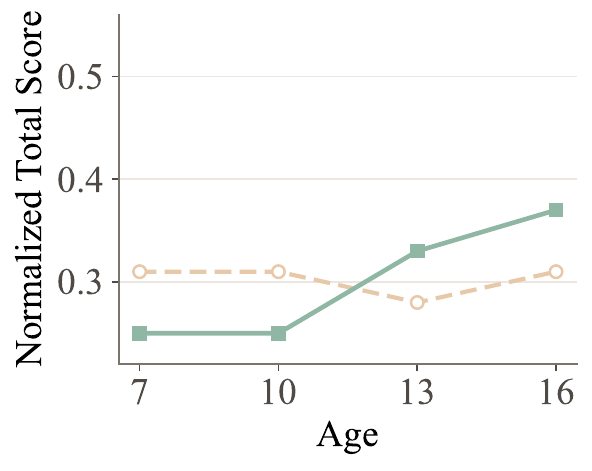}
    \subcaption{Qwen3.6-Plus}
\end{minipage}
\caption{\textbf{Developmental trajectories reveal weak age calibration across proprietary models.}
Each panel reports the normalized total score across the four anchor ages under baseline prompting and skill-guided prompting. }
\label{fig:small_multiples_total}
\end{minipage}
\end{figure*}

The experimental results, detailed in Table 1 and visualized in Fig. \ref{fig:small_multiples_total}, present a clear comparison between the age-agnostic baseline agents and the skill-guided agents.Across highly capable proprietary models such as GPT-5.4, Gemini-3.1-Pro, Gemini-3.1-Flash-Lite and Qwen3.6-Plus, total scores of the baseline agents fail to exhibit a stable age-ordered progression. For instance, the baseline GPT-5.4-based agent scores 0.53 at age 7, drops to 0.46 at age 13, and recovers to 0.52 at age 16. Similar flat or non-monotonic patterns are visible across the other models in the dashed lines of Fig. \ref{fig:small_multiples_total}. This indicates that modifying the nominal age in a system prompt does not meaningfully restrict the reasoning capacity of the model.

\begin{table*}[t]
\centering
\caption{\textbf{Normalized Benchmark Subtest Scores and Age-Normed WISC Composite $z$-Scores for Open-Source Models.}
Total and T1--T10 are normalized benchmark scores; composite columns report age-normed deviations.}
\label{tab:joint_norm_sd_report_open}

\setlength{\tabcolsep}{3.2pt}
\renewcommand{\arraystretch}{1.08}
\scriptsize

\resizebox{\textwidth}{!}{%
\begin{tabular}{llcc cccc ccc cccc ccc}
\toprule
\multirow{2}{*}{\textbf{Method}} &
\multirow{2}{*}{\textbf{Setting}} &
\multicolumn{2}{c}{\textbf{Overall}} &
\multicolumn{4}{c}{\textbf{Gc}} &
\multicolumn{3}{c}{\textbf{WM}} &
\multicolumn{4}{c}{\textbf{Gf/Gv}} &
\multicolumn{3}{c}{\textbf{PSI}} \\
\cmidrule(lr){3-4}
\cmidrule(lr){5-8}
\cmidrule(lr){9-11}
\cmidrule(lr){12-15}
\cmidrule(lr){16-18}
& & \textbf{Total} & \textbf{FSIQ z}
& \textbf{Gc z} & T2 & T6 & T9
& \textbf{WM z} & T3 & T7
& \textbf{Gf/Gv z} & T1 & T4 & T8
& \textbf{PSI z} & T5 & T10 \\
\midrule

\rowcolor{gray!8}
\multicolumn{18}{l}{\textbf{Qwen3.5-27B}} \\
& Baseline (7) & 0.32 & \textbf{0.13} & \textbf{-0.07} & \cellcolor[HTML]{DDEFD8}1.00 & 0.07 & 0.07 & \textbf{5.33} & \cellcolor[HTML]{DDEFD8}1.00 & \cellcolor[HTML]{DDEFD8}1.00 & \textbf{-0.53} & 0.09 & 0.36 & \cellcolor[HTML]{DDEFD8}0.60 & \textbf{-2.93} & 0.02 & 0.15 \\
& Baseline (10) & \cellcolor[HTML]{DDEFD8}0.34 & \textbf{0.00} & \textbf{0.87} & \cellcolor[HTML]{DDEFD8}1.00 & 0.07 & \cellcolor[HTML]{DDEFD8}0.60 & \textbf{5.00} & \cellcolor[HTML]{DDEFD8}1.00 & \cellcolor[HTML]{DDEFD8}1.00 & \textbf{-1.87} & 0.09 & \cellcolor[HTML]{DDEFD8}0.61 & 0.43 & \textbf{-3.67} & 0.02 & 0.02 \\
& Baseline (13) & 0.29 & \textbf{-1.13} & \textbf{-0.93} & 0.95 & 0.09 & 0.14 & \textbf{4.73} & \cellcolor[HTML]{DDEFD8}1.00 & \cellcolor[HTML]{DDEFD8}1.00 & \textbf{-2.80} & 0.09 & 0.29 & 0.40 & \textbf{-3.67} & 0.02 & \cellcolor[HTML]{F6D9D9}0.00 \\
& Baseline (16) & 0.30 & \textbf{-1.00} & \textbf{-1.13} & 0.95 & 0.07 & 0.07 & \textbf{4.47} & \cellcolor[HTML]{DDEFD8}1.00 & \cellcolor[HTML]{DDEFD8}1.00 & \textbf{-2.13} & 0.09 & 0.57 & 0.46 & \textbf{-3.67} & 0.03 & \cellcolor[HTML]{F6D9D9}0.00 \\
\cmidrule(lr){2-18}
& Skill-Guided (6--8) & \cellcolor[HTML]{F6D9D9}0.23 & \textbf{-1.00} & \textbf{-0.07} & 0.82 & 0.07 & 0.07 & \textbf{1.27} & \cellcolor[HTML]{F6D9D9}0.53 & \cellcolor[HTML]{F6D9D9}0.63 & \textbf{-1.53} & 0.09 & 0.36 & 0.31 & \textbf{-2.93} & 0.01 & 0.15 \\
& Skill-Guided (9--11) & \cellcolor[HTML]{F6D9D9}0.23 & \textbf{-1.40} & \textbf{-1.13} & 0.86 & 0.07 & 0.07 & \textbf{2.60} & 0.68 & \cellcolor[HTML]{DDEFD8}1.00 & \textbf{-2.27} & 0.09 & \cellcolor[HTML]{DDEFD8}0.61 & \cellcolor[HTML]{F6D9D9}0.00 & \textbf{-3.67} & \cellcolor[HTML]{F6D9D9}0.00 & \cellcolor[HTML]{F6D9D9}0.00 \\
& Skill-Guided (12--14) & 0.24 & \textbf{-1.73} & \textbf{-1.67} & 0.75 & 0.09 & 0.14 & \textbf{2.87} & 0.79 & \cellcolor[HTML]{DDEFD8}1.00 & \textbf{-2.93} & 0.09 & 0.29 & 0.26 & \textbf{-3.67} & 0.02 & \cellcolor[HTML]{F6D9D9}0.00 \\
& Skill-Guided (15--17) & 0.28 & \textbf{-1.13} & \textbf{-1.13} & 0.95 & 0.07 & 0.07 & \textbf{4.47} & \cellcolor[HTML]{DDEFD8}1.00 & \cellcolor[HTML]{DDEFD8}1.00 & \textbf{-2.40} & 0.09 & 0.57 & 0.31 & \textbf{-3.67} & \cellcolor[HTML]{F6D9D9}0.00 & \cellcolor[HTML]{F6D9D9}0.00 \\

\midrule
\rowcolor{gray!8}
\multicolumn{18}{l}{\textbf{Gemma-4-31B-It}} \\
& Baseline (7) & 0.28 & \textbf{-0.47} & \textbf{0.13} & 0.84 & 0.06 & 0.14 & \textbf{5.33} & \cellcolor[HTML]{DDEFD8}1.00 & 0.97 & \textbf{-2.40} & 0.06 & \cellcolor[HTML]{F6D9D9}0.11 & 0.23 & \textbf{-3.20} & 0.13 & 0.05 \\
& Baseline (10) & 0.29 & \textbf{-1.13} & \textbf{-1.40} & 0.77 & 0.09 & 0.07 & \textbf{5.33} & \cellcolor[HTML]{DDEFD8}1.00 & \cellcolor[HTML]{DDEFD8}1.00 & \textbf{-3.20} & 0.09 & 0.14 & 0.11 & \textbf{-2.73} & \cellcolor[HTML]{DDEFD8}0.14 & \cellcolor[HTML]{DDEFD8}0.20 \\
& Baseline (13) & 0.27 & \textbf{-1.67} & \textbf{-1.60} & 0.82 & 0.07 & 0.12 & \textbf{4.47} & \cellcolor[HTML]{DDEFD8}1.00 & 0.93 & \textbf{-3.60} & \cellcolor[HTML]{F6D9D9}0.00 & 0.14 & 0.17 & \textbf{-3.67} & 0.12 & 0.10 \\
& Baseline (16) & 0.25 & \textbf{-1.73} & \textbf{-1.67} & 0.82 & \cellcolor[HTML]{F6D9D9}0.04 & 0.12 & \textbf{3.93} & \cellcolor[HTML]{DDEFD8}1.00 & 0.93 & \textbf{-3.47} & \cellcolor[HTML]{F6D9D9}0.00 & 0.14 & 0.26 & \textbf{-3.67} & 0.05 & 0.07 \\
\cmidrule(lr){2-18}
& Skill-Guided (6--8) & \cellcolor[HTML]{F6D9D9}0.23 & \textbf{-0.87} & \textbf{-1.13} & \cellcolor[HTML]{F6D9D9}0.61 & 0.06 & \cellcolor[HTML]{F6D9D9}0.02 & \textbf{5.33} & \cellcolor[HTML]{DDEFD8}1.00 & \cellcolor[HTML]{DDEFD8}1.00 & \textbf{-2.27} & 0.06 & 0.14 & 0.26 & \textbf{-3.40} & 0.06 & 0.02 \\
& Skill-Guided (9--11) & 0.25 & \textbf{-1.73} & \textbf{-1.27} & 0.80 & \cellcolor[HTML]{DDEFD8}0.10 & 0.10 & \textbf{1.93} & \cellcolor[HTML]{F6D9D9}0.53 & \cellcolor[HTML]{DDEFD8}1.00 & \textbf{-3.07} & 0.09 & 0.14 & 0.23 & \textbf{-3.40} & 0.08 & 0.13 \\
& Skill-Guided (12--14) & 0.27 & \textbf{-1.53} & \textbf{-1.47} & 0.84 & 0.09 & 0.10 & \textbf{4.73} & \cellcolor[HTML]{DDEFD8}1.00 & \cellcolor[HTML]{DDEFD8}1.00 & \textbf{-3.33} & \cellcolor[HTML]{DDEFD8}0.15 & 0.14 & 0.06 & \textbf{-3.67} & 0.08 & 0.03 \\
& Skill-Guided (15--17) & 0.27 & \textbf{-1.67} & \textbf{-1.67} & 0.82 & 0.09 & 0.07 & \textbf{4.47} & \cellcolor[HTML]{DDEFD8}1.00 & \cellcolor[HTML]{DDEFD8}1.00 & \textbf{-3.60} & 0.09 & 0.14 & 0.09 & \textbf{-3.67} & 0.08 & 0.10 \\
\bottomrule
\end{tabular}%
}

\par\vspace{2pt}
\begin{minipage}{0.98\textwidth}
\footnotesize
\textit{Note.}
T denotes test index. `Total` and `T1--T10` report normalized benchmark scores derived directly from the web-based task outcomes. By contrast, `FSIQ z`, `Gc z`, `WM z`, `Gf/Gv z`, and `PSI z` report age-referenced deviation scores computed from WISC-style composite scores after age-based norm conversion. Specifically, raw subtest scores are first mapped to age-referenced scaled scores, these scaled scores are then aggregated into composite scores, and the reported deviation values are finally computed as $z=(S-100)/15$, where $S$ denotes the corresponding composite score on the WISC-style normative scale. For the skill-guided condition, the four rows correspond to target age bands 6--8, 9--11, 12--14, and 15--17. Soft green marks the highest value and soft red marks the lowest value among non-$z$ numeric columns within each table block.

\end{minipage}
\end{table*}

When the cognitive skill distillation is applied, the developmental trajectories shift significantly. As shown in the solid lines of Fig. \ref{fig:small_multiples_total}, the skill-guided condition induces a monotonic increase in total scores from age 7 to age 16 for all evaluated proprietary models. For GPT-5.4-based agent, the total score scales consistently from 0.41 at the 6-8 age band to 0.50 at the 15-17 age band. The consistency of this trend across several proprietary architectures suggests that the proposed cognitive filters can induce age-ordered differentiation in sufficiently capable models. 

However, the results also expose a capability threshold required for cognitive simulation. This suggests that the current constraint design does not yet generalize uniformly across model families, and instead depends on a sufficiently high level of baseline controllability and instruction-following capacity.
 While proprietary models demonstrate general compliance with the constraints, both Qwen3.5-27B-based agent and Gemma-4-31B-It-based agent remain comparatively weak and do not exhibit the clear age-ordered trajectories. Under the skill-guided setting, their scores change only modestly across ages, indicating limited calibration rather than stable developmental alignment. If the base model lacks this capacity, the constraints cause task failure rather than age calibration.

Ultimately, these main results indicate a shift in evaluation goals. In traditional agent benchmarks, lower scores indicate failure. In ChildAgentEval, the fact that a 7-year-old calibrated agent scores significantly lower than its baseline counterpart is an indicator of successful alignment, provided that performance expands in an age-ordered manner as the target age increases. The models do not simply answer randomly; they demonstrate bounded reasoning that expands as the target age increases. Specifically, this exposes a factor-specific mismatch against child norms: while Gc and language-related dimensions exhibit clear age-ordered scaling, Gf/Gv, WMI and PSI remain far less sensitive to developmental constraints. Detailed factor-level trajectories and cross-model profiles supporting these observations are provided in Appendix~\ref{app:factor_trajectories_appendix} and Appendix~\ref{app:radar_profiles_appendix}. These findings show that skill-guided age alignment is both achievable and measurable. To mathematically substantiate these overarching observations, the following section transitions from qualitative trends to a rigorous statistical breakdown.

\subsection{Quantitative and Age-Normed Analyses}

\begin{figure*}[t]
\centering
\begin{minipage}{0.98\textwidth}
\centering
\captionsetup[subfigure]{labelfont={times}}
 
\begin{minipage}[t]{0.32\textwidth}
    \centering
    \includegraphics[width=\linewidth]{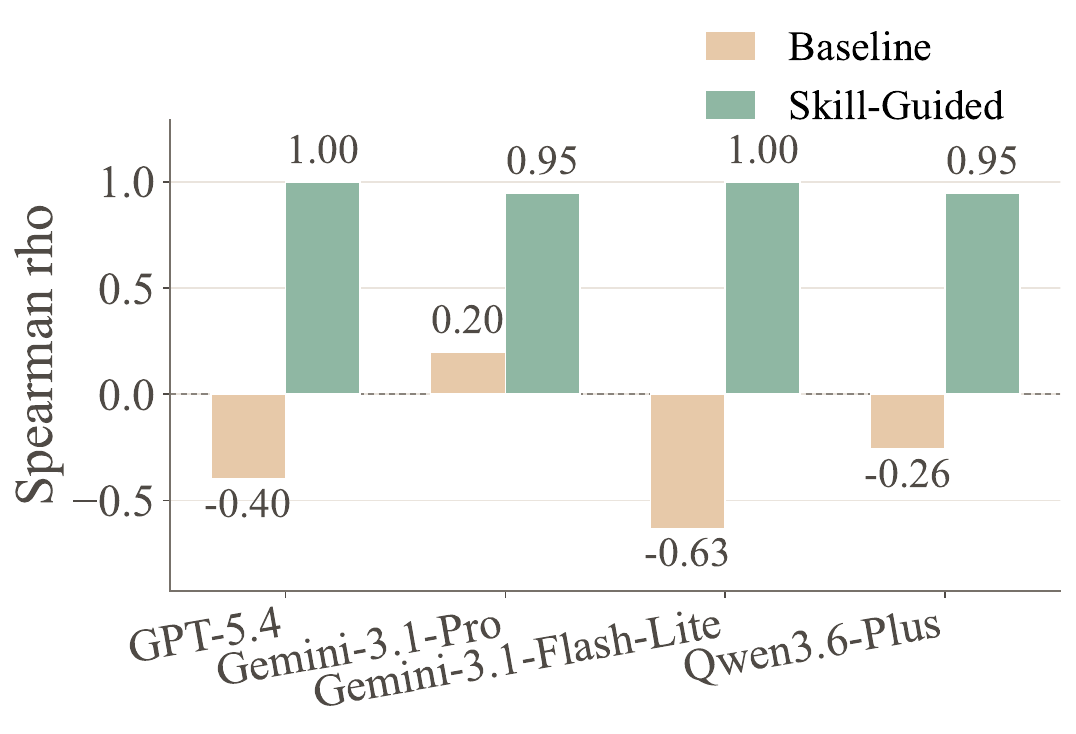}
    \subcaption{Spearman $\rho$}
\end{minipage}
\hfill
\begin{minipage}[t]{0.32\textwidth}
    \centering
    \includegraphics[width=\linewidth]{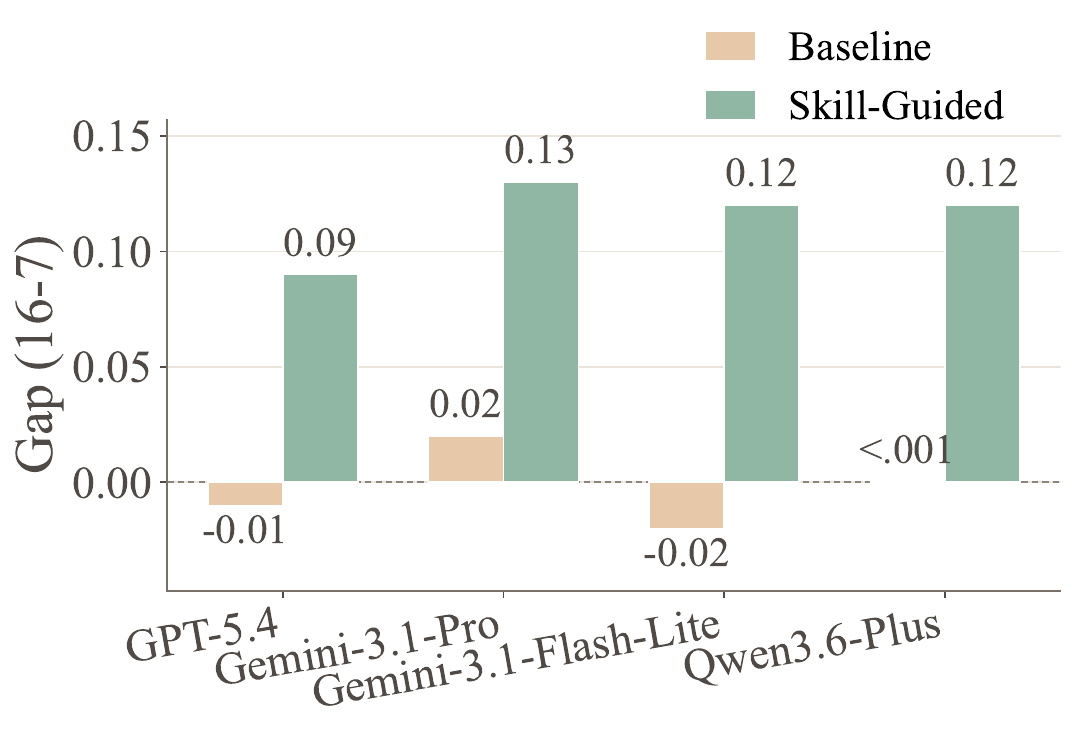}
    \subcaption{Gap (16--7)}
\end{minipage}
\hfill
\begin{minipage}[t]{0.32\textwidth}
    \centering
    \includegraphics[width=\linewidth]{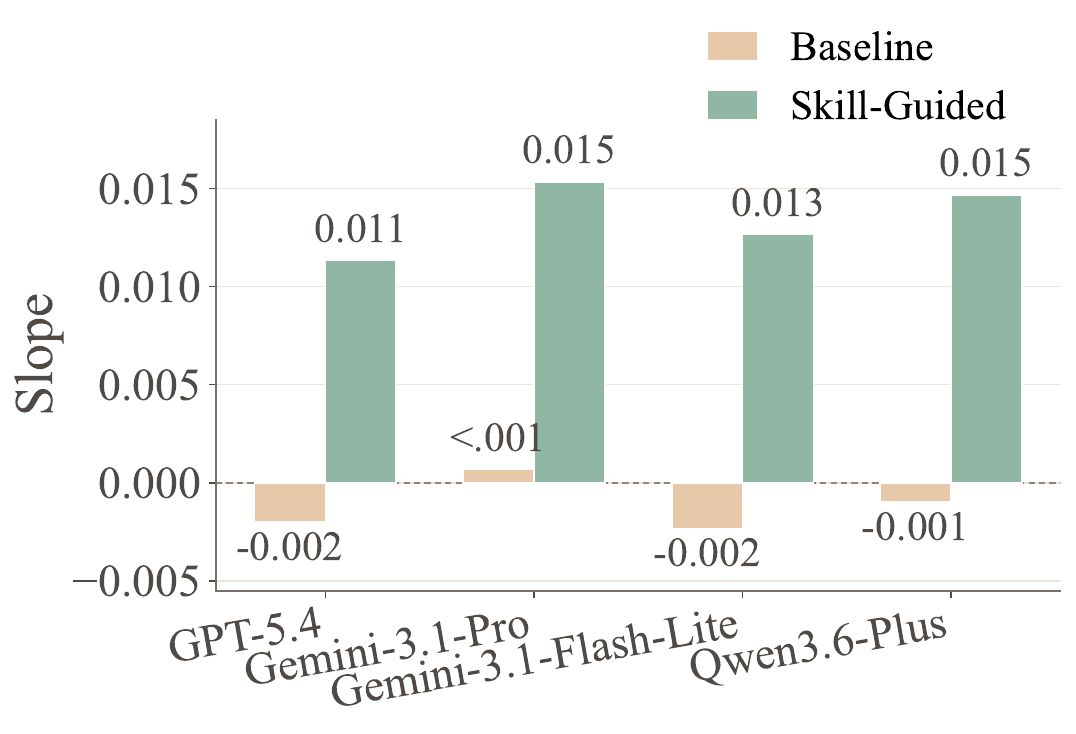}
    \subcaption{Slope}
\end{minipage}
\caption{\textbf{Skill guidance increases developmental differentiation across models.}
We compare baseline and skill-guided settings using four trajectory-level metrics: (a) Spearman rank correlation between target age and total score, (b) the total-score gap between ages 16 and 7 and (c) the regression slope across the four anchor ages. }
\label{fig:dev_metrics}
\end{minipage}
\end{figure*}

\paragraph{Quantitative Analysis of Developmental Differentiation.}
To rigorously evaluate whether the cognitive skill constraints induce true developmental differentiation, we extract statistical metrics from the normalized score trajectories as visualized in Fig. \ref{fig:dev_metrics}. The baseline configurations across all models fail to produce meaningful developmental progression. For example, the baseline GPT-5.4-based agent yields a negative Spearman rank correlation of -0.40 and a negative score gap of -0.01 between age 16 and age 7. Similar negative or near-zero correlations are observed for Gemini-3.1-Flash-Lite-based agent and Qwen3.6-Plus-based agent under baseline settings. 


Conversely, the skill-guided setting substantially changes the developmental trajectory. 
For GPT-5.4, total performance increases monotonically with the target age, yielding a Spearman correlation of 1.00 and increasing the age-16 versus age-7 gap to 0.09. Although the absolute range remains modest, the monotonicity of this shift is consistent across the stronger proprietary models.
These shifts suggest that skill guidance does more than lower overall accuracy: it reshapes the model's response behavior so that performance varies more consistently with the intended developmental level.
This result also clarifies that  approximating child-like developmental profiles appears to require targeted constraints on the model's reasoning process, rather than merely asking the model to ``act younger.''\looseness-1

\paragraph{Age-Normed Deviation Profiles.} Age-normed $z$-scores (Tables~\ref{tab:joint_norm_sd_report_prop} and~\ref{tab:joint_norm_sd_report_open}) reveal a structured mismatch between MLLM strengths and human developmental norms. Gemini-3.1-Pro exemplifies this: its Gc and WM indices are positively shifted, indicating that language-mediated reasoning and short-term retention exceed age-specific expectations. Conversely, its Gf/Gv index aligns only at the youngest anchor before failing to keep pace with normative trajectories, while PSI remains consistently below the norm, highlighting persistent weaknesses in speed-dependent visual-symbolic performance.  

This domain dissociation also characterizes GPT-5.4 and Gemini-3.1-Flash-Lite, though Gemini-3.1-Pro more clearly illustrates a reasoning profile that begins age-matched but progressively regresses. In contrast, open-weight models like Qwen and Gemma generally underperform across multiple factors simultaneously. Ultimately, $z$-score analysis indicates that MLLMs do not drift uniformly from child norms; instead, they preserve disproportionately strong language and memory capacities while remaining systematically weak in the growth of perceptual reasoning and processing speed. 

\section{Discussion}
\label{sec:discussion}
Cognitive age alignment requires selective behavioral reconfiguration rather than uniform capability reduction. Standard prompting fails to elicit age-ordered trends because agents prioritize task correctness over behavioral consistency. While skill guidance improves calibration by constraining reasoning, memory, and vocabulary, alignment remains uneven. MLLMs easily adapt linguistic style, but architectural bottlenecks prevent the authentic reproduction of human-like memory decay or perceptual limits. Consequently, current agents primarily imitate child-like surface features while retaining adult-level cognitive structures. In sensitive applications like educational tutoring, developmental appropriateness must supersede raw accuracy. \benchmarknamenc{} shifts evaluation to prioritize this cognitive alignment. Future research should explore age-specific post-training to embed developmental constraints directly into the model's core architecture, bridging the gap between surface mimicry and structural alignment.

\section{Conclusion}
\vspace{-1mm}
This work introduces \benchmarknamenc{}, an interactive, WISC-grounded framework and a data-driven skill distillation method to evaluate and implement developmental alignment in MLLM agents. We demonstrate that nominal age instructions are insufficient, as general-purpose agents default to their maximum capabilities. However, targeted cognitive filters enable monotonic score progressions and age-ordered linguistic patterns in high-performing models. Our findings establish that authentic alignment requires more than stylistic role-play; it necessitates fundamental constraints on the cognitive processes of perception, memory, and reasoning.

\clearpage
\bibliography{main}
\bibliographystyle{bibstyle}

\newpage
\appendix
\section*{Appendix Contents}

\startcontents[appendices]
\printcontents[appendices]{}{1}{}

\newpage



\section{More Implementation Details}
\label{app:implement}

\subsection{Execution modes: vision-only vs. DOM-assisted.}
We implemented two interaction modes for the browser environment: \textit{vision-only} and \textit{DOM-assisted}. In the \textit{vision-only} mode, the agent relies entirely on rendered screenshots to formulate and execute browser actions (e.g., selecting, typing). In the \textit{DOM-assisted} mode, the agent receives the screenshot alongside a sanitized accessibility tree detailing the visible interactive elements, their roles, and bounding boxes. Crucially, this DOM summary is strictly filtered to exclude hidden states, answer keys, and backend data attributes. It serves solely to facilitate spatial action grounding without providing cognitive shortcuts. To maintain testing validity, both modes require the agent to execute interactions via Playwright; direct textual responses to the evaluator are prohibited. We adopt the \textit{DOM-assisted} mode for our primary experiments to isolate cognitive capabilities from confounding visual localization errors. This design ensures that measured failures reflect genuine reasoning deficits rather than basic pixel-level misalignments, while also improving the reproducibility of the action logs.

\subsection{Calibration Data and Separation from Evaluation}

The cognitive filters are calibrated using external developmental data and markers, rather than by directly fitting the benchmark evaluation scores. Specifically, the skill configurations are derived from age-stratified corpora and developmental summaries. These sources define the target thresholds for vocabulary abstraction, memory capacity, reasoning depth, and related behavioral constraints. 

To prevent data leakage, the benchmark evaluation items are not used as supervision targets for tuning these filters. We do not optimize the skill configurations to match the final benchmark score tables, nor do we use benchmark answer keys or item-level scores as a training objective. Therefore, the calibration stage is designed to specify developmentally motivated constraints rather than to numerically fit the model to the benchmark.

Consequently, the current pipeline operates as a constraint-design procedure grounded in developmental evidence, not as a score-matching procedure on the evaluation set. Future work will further validate this separation by utilizing larger held-out calibration corpora and conducting formal ablation studies on the transferability of individual filters.

\subsection{Scoring and human verification.}
Objective subtests are scored deterministically from the browser state after the agent submits an answer. For exact-match tasks, such as Digit Span and Letter--Number Sequencing, the submitted string must match the target sequence after standard normalization. For selection tasks, such as Picture Concepts and Matrix Reasoning, the selected option set must match the ground truth. For timed
processing-speed tasks, the raw score is the number of correct operations completed within the time limit.

Open-ended verbal items require human judgment. This includes Similarities, advanced Vocabulary items, and Comprehension. GPT-5.4 is used only as a pre-annotation assistant to generate a tentative score and flag potentially ambiguous cases. No GPT-only score is used in the final reported results. All open-ended responses are anonymized and verified by human raters who are blind to the model identity, experimental setting, target age, and trajectory-level hypothesis. The raters see only the item prompt, the scoring rubric, and the agent response. Any explicit metadata accidentally produced by the agent, such as phrases revealing the prompted age or model identity, is removed before grading when it is not part of the substantive answer.

Each open-ended response is independently scored by two human raters using the standard 0/1/2 rubric: 2 points for a complete and abstractly appropriate answer, 1 point for a partially correct or overly concrete answer, and 0 points for an incorrect, irrelevant, or missing answer. If the two raters agree, their shared score is used as the final score. If they disagree, the item is adjudicated by a third reviewer or a child-psychology-trained annotator. The final benchmark tables use only the human-verified scores.

\subsection{Computational Execution and Reproducibility}
\label{time execution}

To ensure deterministic evaluation, proprietary models are accessed through their official APIs using greedy decoding with a temperature of \(0.0\). We parallelize the evaluation at the administration level, where each worker controls an independent Playwright browser context to execute one specific model, setting, and age configuration at a time. A complete administration requires approximately one hour of wall-clock time. This duration naturally varies depending on API latency, model response length, browser execution time, and the exact number of items administered before the discontinuation rule is triggered. Furthermore, timed subtests such as Coding and Symbol Search introduce fixed waiting periods that cannot be skipped, because these strict time limits are integral to the cognitive construct being measured.

To guarantee complete reproducibility, the platform records extensive artifacts for every evaluation run. The system stores the model configuration, target age band, experimental setting, item-level submissions, raw scores, and converted scaled scores. Alongside the assessment data, it saves detailed browser action traces, page-transition logs, and timing statistics. These comprehensive telemetry logs allow us to systematically audit how every final score is calculated and to explicitly separate genuine cognitive errors from underlying interface or infrastructure failures.

\subsection{Statistical Framing of the Main Results}

The analyses presented in the main text focus on the descriptive evaluation of developmental trajectories rather than inferential statistical testing. Because all experiments are conducted using greedy decoding and fixed administration protocols, the resulting age profiles characterize structured behavioral responses under strictly controlled conditions, rather than stochastic outcome distributions over repeated random trials. 

Consequently, the trajectory-level metrics reported, including the Spearman rank correlation, age-gap scores, and regression slopes, serve as deterministic summaries of age-ordered behavioral differentiation. The current evaluation framework does not compute confidence intervals, formal significance tests, or variance estimates derived from repeated runs at the subtest or factor level. Future research can expand upon this benchmark by introducing stochastic sampling, bootstrap confidence intervals, and formal inferential comparisons across different models and cognitive constraints.

\subsection{Use of WISC Normative Scoring}
\benchmarknamenc{} compares agents with human developmental performance through age-stratified normative scoring. In psychometric evaluation, the human baseline is not usually a newly recruited small control group, but the official normative table constructed from large samples of children. We follow this convention. For each target age, raw subtest scores are converted using age-specific normative mappings. The official age-stratified norms provide a large-scale and standardized human reference, while the normalized raw scores provide within-benchmark comparisons among agents. We use norm-referenced scoring as the primary human baseline

\section{Factor-Level Analysis and Multidimensional Profiles}
\label{analysis}
\subsection{Fine-grained Factor-Level Trajectories}
\label{app:factor_trajectories_appendix}

\begin{figure*}[t]
\centering
\begin{minipage}{0.98\textwidth}
\centering
\captionsetup[subfigure]{labelfont={times}}
\begin{minipage}[t]{0.24\textwidth}
    \centering
    \includegraphics[width=\linewidth]{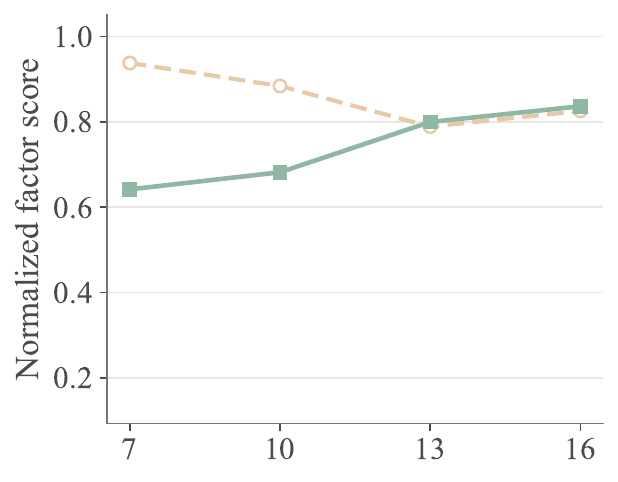}
    \subcaption{Gc}
\end{minipage}
\hfill
\begin{minipage}[t]{0.24\textwidth}
    \centering
    \includegraphics[width=\linewidth]{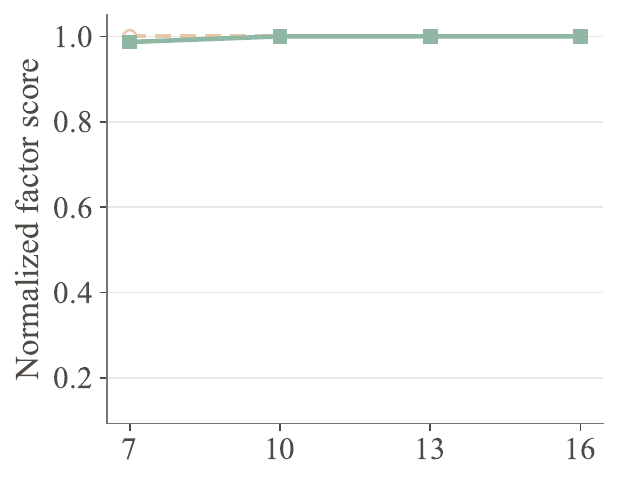}
    \subcaption{WM}
\end{minipage}
\hfill
\begin{minipage}[t]{0.24\textwidth}
    \centering
    \includegraphics[width=\linewidth]{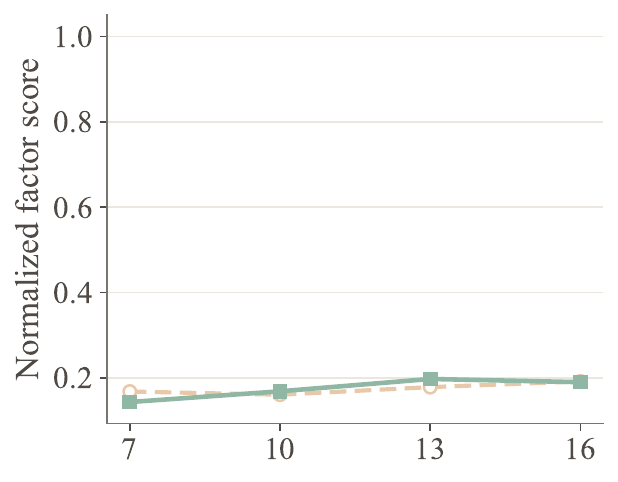}
    \subcaption{Gf/Gv}
\end{minipage}
\hfill
\begin{minipage}[t]{0.24\textwidth}
    \centering
    \includegraphics[width=\linewidth]{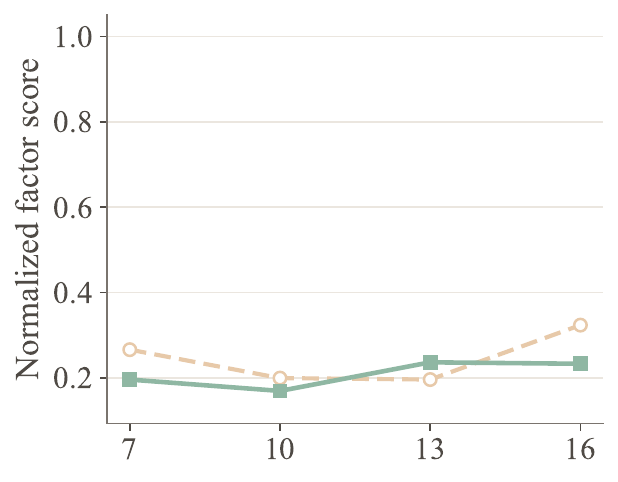}
    \subcaption{PSI}
\end{minipage}
\caption{\textbf{Factor-level developmental trajectories for GPT-5.4 under the baseline and skill-guided settings.} Panels show normalized scores for Gc, WM, Gf/Gv and PSI across 4 anchor ages.}
\label{fig:factor_trajectories}
\end{minipage}
\end{figure*}

While the total scores provide a macroscopic view, we must decompose the performance into specific cognitive dimensions to understand the mechanics of the degradation. Fig. \ref{fig:factor_trajectories} illustrates the developmental trajectories of GPT-5.4 across four recognized psychometric factors. 

The baseline trajectories remain largely flat, non-monotonic, or pegged at performance extremes across all factors. When the cognitive constraints are applied, only the Gc factor demonstrates a clear, monotonic upward scaling. Specifically, the skill-guided Gc score for the GPT-5.4-based agent rises from 0.64 at age 7 to 0.84 at age 16. This targeted suppression at lower ages indicates that the vocabulary boundaries and abstractness filters are functioning correctly. In contrast, the Gf/Gv factor shows only a marginal trajectory improvement, remaining largely flat and close to its baseline. This suggests a potential floor effect, indicating that complex fluid reasoning and spatial manipulation tasks are inherently difficult for the model or are less responsive to current prompt-based developmental constraints, which is consistent with recent evidence of a broader visual cognition gap in multimodal LLMs~\citep{cao2024visual}.\looseness-1

Furthermore, WM and PSI exhibit distinct insensitivities to the simulated age settings. WM remains entirely saturated at the performance ceiling across the entire simulated age span, while PSI fluctuates at a lower performance tier without establishing a clear developmental trend. We hypothesize that these discrepancies arise from the architectural nature of large language models. While semantic knowledge (Gc) can be effectively restricted through tailored instruction sets, working memory is inherently tied to the fixed context window, and processing speed is governed by the rigid computational graph of the neural network. Modifying these structural properties through prompt-based filters is difficult. This finding suggests that future research must directly limit the attention mechanism or token retention algorithms across dialogue turns to accurately simulate the limited biological working memory and processing bottlenecks of a child.

\subsection{Linguistic Profiles and Cross-Model Cognitive.}
\label{app:radar_profiles_appendix}

To provide an additional layer of analysis alongside the cognitive profiles, we define a supplementary language-complexity dimension, denoted as $Lang.$, for the open-ended verbal subtests. This analysis is computed from the generated responses in three open-ended subtests: Similarities (T2), Vocabulary (T6), and Comprehension (T9). The objective of this dimension is to quantify whether age calibration induces systematic shifts in response length, lexical diversity, abstraction, categorization, and explanatory structure.
Specifically, $Lang.$ is computed from seven component metrics: mean length of utterance (MLU)~\citep{brown1973firstlanguage}, moving-average type-token ratio (MATTR)~\citep{covington2010mattr}, abstractness~\citep{brysbaert2014concreteness}, category rate, causal rate, definition rate, and average number of reasons. For each metric $x_k$, we apply max-normalization over the evaluated model conditions and then compute the arithmetic mean $Lang.\!=\!\frac{1}{7} \sum_{k=1}^{7} \frac{x_k}{\max_{\mathcal{M}}(x_k)}$,
where $\mathcal{M}$ denotes the set of evaluated model conditions in the current analysis.\looseness-1

Among these components, MLU serves as a baseline indicator of response length, while MATTR provides a length-robust measure of lexical diversity. English responses are tokenized using standard alphabetic word segmentation, whereas Chinese responses are tokenized at the character level. MLU is calculated as the total number of resulting tokens per response. MATTR is computed over the same token sequence utilizing a sliding window of 50 units; for responses shorter than 50 tokens, the metric defaults to the standard type-token ratio of the entire sequence.

The remaining five components are implemented as response-level binary or frequency features. Abstractness is a binary indicator identifying whether a response contains at least one abstract lexical item from a pre-defined lexicon. Category rate is a binary indicator marking the presence of superordinate categorization markers, such as ``type of'' or equivalent structures. Causal rate and definition rate are similarly defined as binary indicators for the presence of causal connectives and definitional syntactic patterns (e.g., ``is '', ``means''), respectively. The average number of reasons is computed by counting the frequency of causal markers within each response and averaging across the dataset. We note that this specific metric operates as a proxy for structural explanatory density, rather than a semantic count of distinct logical arguments.

For each model, condition, and age group, the resulting per-age values construct the final language-complexity composite. Given that these linguistic indicators share the same source material as the open-ended subtests comprising the Gc factor, we present $Lang.$ strictly as an analysis of structural response form, acknowledging the shared variance between linguistic complexity and crystallized intelligence.

\begin{figure*}[t]
\centering
\begin{minipage}{0.98\textwidth}
\centering
 \captionsetup[subfigure]{labelfont={times}}

\begin{minipage}[t]{0.24\textwidth}
    \centering
    \includegraphics[width=\linewidth]{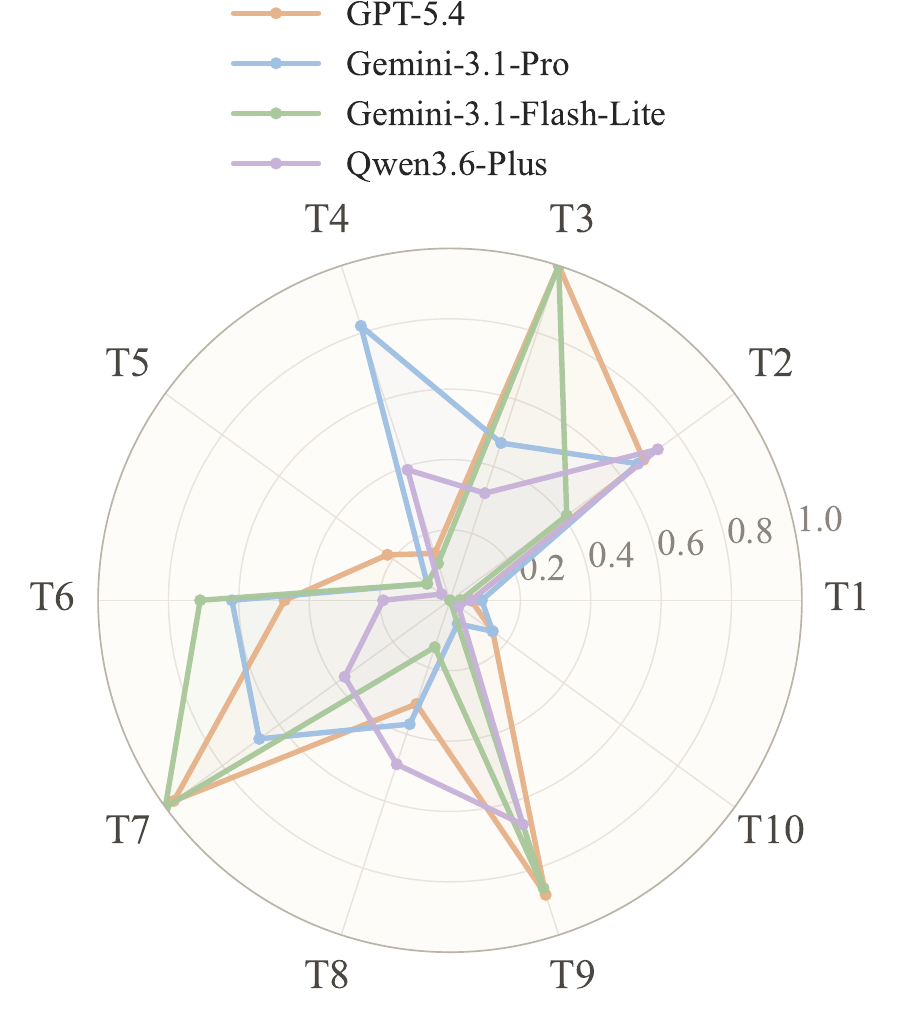}
    \vspace{-7mm}
    \subcaption{Age 7 subtests}
\end{minipage}
\hfill
\begin{minipage}[t]{0.24\textwidth}
    \centering
    \includegraphics[width=\linewidth]{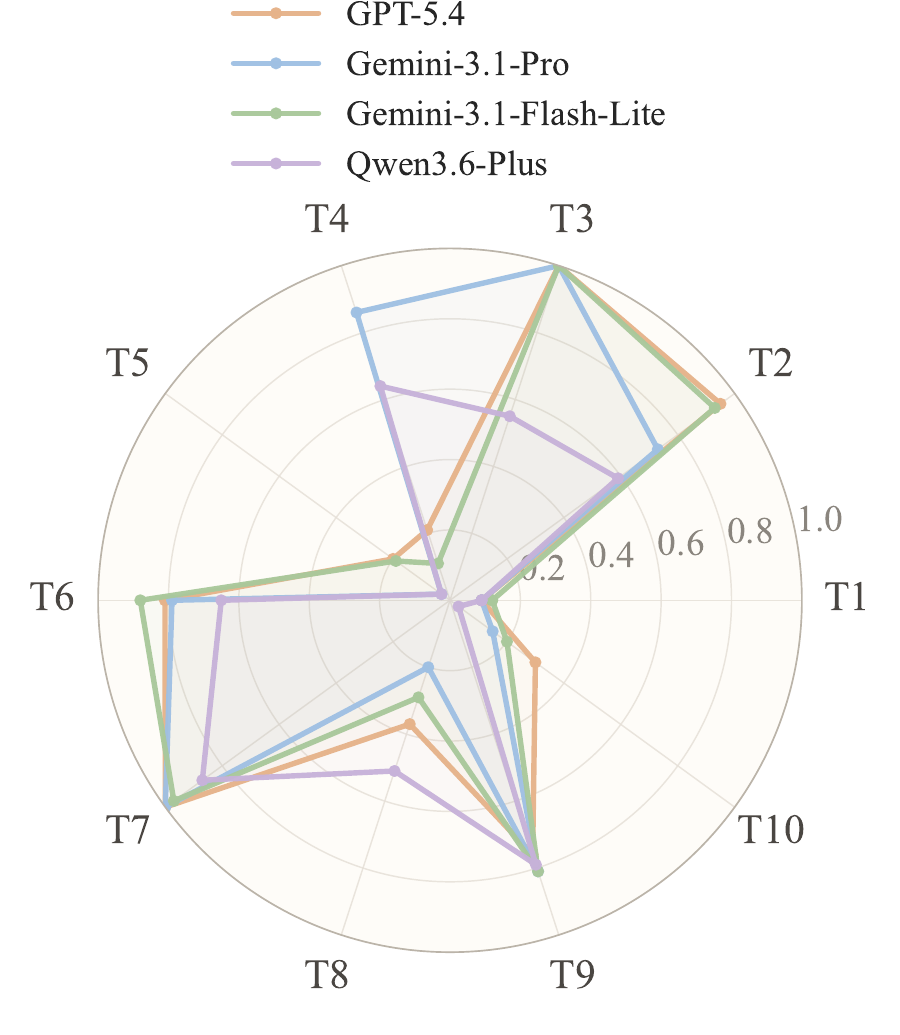}
    \subcaption{Age 16 subtests}
\end{minipage}
\hfill
\begin{minipage}[t]{0.24\textwidth}
    \centering
    \includegraphics[width=\linewidth]{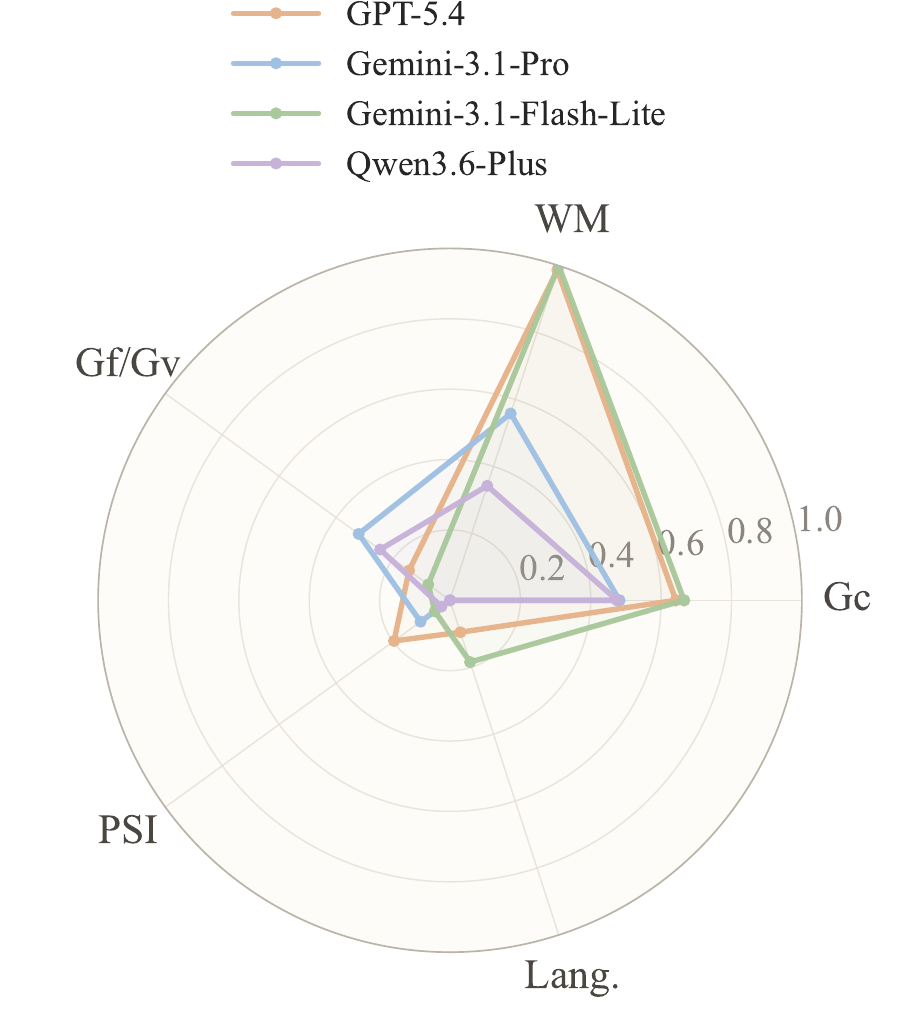}
    \subcaption{Age 7 factors}
\end{minipage}
\hfill
\begin{minipage}[t]{0.24\textwidth}
    \centering
    \includegraphics[width=\linewidth]{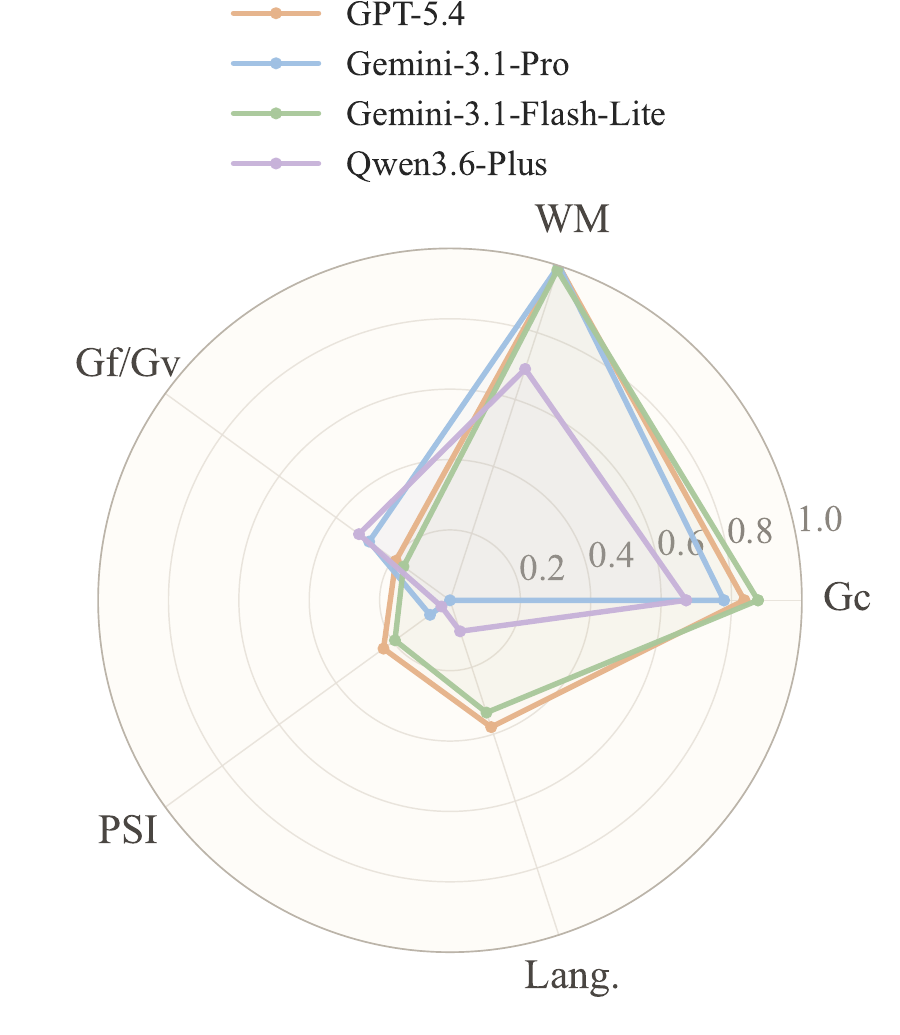}
    \subcaption{Age 16 factors}
\end{minipage}
\caption{\textbf{Skill-guided age simulation produces distinct cognitive profiles across models.}
Panels (a) and (b) compare normalized subtest-level radar profiles at ages 7 and 16, respectively. 
Panels (c) and (d) aggregate these patterns into factor-level profiles and include language complexity, showing how cognitive performance and linguistic behavior jointly vary across models and target ages.}
\label{fig:radar_profiles}
\end{minipage}

\end{figure*}

Next, we analyzed the multidimensional capability profiles induced by age calibration to assess the balance of agent performance across specific subtests and cognitive factors and linguistic behavior. Fig. \ref{fig:radar_profiles} presents multi-dimensional radar visualizations of the resulting profiles. Panels (a) and (b) of Fig. \ref{fig:radar_profiles} map the performance across the ten individual subtests, contrasting the simulation at age 7 against age 16. The expansion of the polygon area from the younger to the older simulation visually represents the calibrated release of reasoning capabilities. This expansion is most visible in verbally mediated subtests and in several reasoning-related tasks, whereas speeded tasks remain relatively compressed even at the older anchor. The resulting geometry therefore reveals not only overall growth, but also which abilities remain disproportionately weak or strong under calibration.

To verify that these changes are not limited to test success rates, panels (c) and (d) of Fig. \ref{fig:radar_profiles} integrate ($Lang.$) alongside the four cognitive factors. The data show that language complexity scales synchronously with cognitive capacity. For the GPT-5.4-based agent, the language composite score expands from 0.09 at age 7 to 0.38 at age 16. Here, the main pattern is not simply that older simulated agents score higher, but that the balance among factors changes in a structured way. For example, GPT-5.4-based and Gemini-3.1-Flash-Lite-based agent show a clear outward shift from age 7 to age 16 in both Gc and the language dimension, indicating that stronger verbal knowledge is accompanied by more elaborate linguistic output. By contrast, Gf/Gv and especially PSI expand less dramatically, suggesting that the developmental release induced by skill conditioning is not uniform across cognitive domains. Qwen3.6-Plus also shows some age-related expansion, but the overall profile remains more compressed, indicating weaker differentiation under the same constraints.

Importantly, these specific improvements at the subtest-level provide the underlying explanation for the broader factor-level shifts shown in the companion panels. As performance on language-related and verbally mediated tasks increases, it directly drives the outward expansion of higher-order dimensions like Gc and $Lang.$ We consider this synchronous growth between cognitive capacity and language complexity to be a critical scientific finding. It demonstrates that the models constrained to simulate younger ages do not achieve lower scores simply by randomly guessing or artificially failing tests. Rather, their restricted internal reasoning organically yields simpler, more concrete external language. Ultimately, this tight alignment between cognitive problem-solving logic and linguistic behavior provides compelling evidence that our skill distillation method accurately simulates the integrated behavioral patterns of human cognitive development.\looseness-1


\section{Data Collection and Processing Details}
\label{app:data_details}

To accurately capture the cognitive features of different developmental stages, we assembled and integrated a multi-source corpus covering ages 6 to 17. For the lower age group of 6 to 11 years old, we mainly used spoken and multimodal interaction data such as CHILDES~\citep{e8742e7c7f334a54b4ded29732ff174c}, OCSC~\citep{WAGNER2025103206}, and Frog Story~\citep{REILLY2004229}. These spoken data can effectively reflect the daily vocabulary boundaries, immediate attention spans, and self-repair markers in natural conversations of children. 

For the higher age group of 12 to 17 years old, we introduced corpora such as LCCPW~\citep{5ba219b09aa243438842d1cc4cd199c6} and ClassBank~\citep{NesiForthcoming-NESIEO, al2025classbank}, focusing on extracting classroom discussions, psychological interviews, and narrative writing texts. Writing and interview data provide data support for the use of abstract vocabulary, the organization of long-range logical reasoning, and the egocentric bias specific to adolescents. 

In the data processing stage, we strictly divided all corpora into four target age groups based on metadata: 6 to 8 years old, 9 to 11 years old, 12 to 14 years old, and 15 to 17 years old. To evaluate lexical diversity, we applied the MATTR framework, which calculates the moving average for the ratio of types to tokens~\citep{covington2010mattr}. To extract structural and fluency metrics, including Mean Length of Utterance, grammatical depth, and mid-course corrections, we used the standard CLAN toolkits provided by TalkBank. Finally, we inputted these extracted statistical distributions and sampled corpus fragments into a teacher language model (GPT-5.4) to distill the raw linguistic features into structured, age-specific cognitive constraints.

\section{Role of \benchmarknamenc{}}
\benchmarknamenc{} functions as an evaluation infrastructure rather than a static benchmark dataset. It provides a standardized, web-based environment to administer cognitive tasks to agents under consistent interaction and scoring protocols. Because the system is strictly model-agnostic, external researchers can seamlessly integrate their own large language models into the platform to be evaluated as interactive agents. To conduct an evaluation, users apply the Age-Specific Cognitive Skill Distillation resource, which provides the necessary parameters to configure any target model for specific developmental stages. This architecture establishes the platform as a universal execution layer, while the distillation resource defines the cognitive constraints required for age-aligned assessment independent of the underlying model.

\section{Broader Impacts}
\benchmarknamenc{} may support the development of safer and more developmentally appropriate child-facing AI agents, especially in education, tutoring, and assistive interaction scenarios. By evaluating whether agents align their language, reasoning, memory use, and explanation style with target developmental stages, the benchmark encourages evaluation beyond raw task accuracy. Potential risks include over-interpreting benchmark scores as clinical measurements or misusing age-simulation ability for deceptive child-like impersonation. To reduce these risks, the benchmark is intended only for research use, does not reproduce protected clinical test items, and should be applied with human oversight and expert review in child-facing settings.

\section{Limitations}

While the benchmark maps human cognitive factors to agent behaviors, specific evaluations must be interpreted as operational simulations rather than biological reconstructions. For processing speed, the time-constrained browser tasks measure overall pipeline efficiency. This efficiency is inherently affected by system-level factors, including model inference time, API response latency, and tool orchestration overhead. Consequently, even under identical task protocols, we cannot completely isolate pure cognitive execution speed from these backend delays. Similarly, the working memory control mechanism operates as a functional approximation. We regulate memory demands through interface design, progressive presentation, and context separation to prevent the model from exploiting its full context window. While this approach successfully approximates age-dependent limits on short-term retention, it acts as an external constraint on accessible information. It does not alter the intrinsic architecture of the model to replicate human memory decay or biological attentional bottlenecks. Ultimately, both processing speed and working memory in this framework represent constrained operational metrics rather than direct structural implementations of human cognitive limits.

This work is intended as a first step toward evaluating cognitive age alignment in interactive AI agents. The current benchmark is implemented in a controlled browser-based environment, which enables standardized administration and detailed logging but leaves other interaction formats, such as voice-based or long-horizon tutoring scenarios, to future work. We also evaluate several representative age bands to obtain stable developmental trajectories; future extensions could examine finer-grained age intervals. Finally, although \benchmarknamenc{} records rich interaction telemetry, the present paper mainly analyzes score trajectories and factor-level profiles. More detailed process-level modeling of action timing and navigation behavior is an interesting direction for future study.

\end{document}